\pdfoutput=1
\documentclass[10pt,twocolumn,letterpaper]{article}

\usepackage[pagenumbers]{cvpr}

\usepackage{graphicx}
\usepackage{subcaption}

\usepackage{cuted} 

\usepackage{currfile} 

\usepackage{caption} 

\newcommand{\TODO}[1]{\textbf{\color{red}[TODO: #1]}}

\usepackage{microtype}

\renewcommand{\paragraph}[1]{\vspace{.5em}\noindent\textbf{#1.}}

\AtBeginDocument{%
  \setlength{\abovedisplayskip}{6pt plus 2pt minus 2pt}%
  \setlength{\belowdisplayskip}{6pt plus 2pt minus 2pt}%
  \setlength{\abovedisplayshortskip}{3pt plus 1pt minus 1pt}%
  \setlength{\belowdisplayshortskip}{3pt plus 1pt minus 1pt}%
  \setlength{\jot}{2pt}%
}

\AtBeginDocument{%
  \setlength{\textfloatsep}{10pt plus 2pt minus 2pt}
  \setlength{\dbltextfloatsep}{10pt plus 2pt minus 2pt}
  \setlength{\floatsep}{8pt plus 2pt minus 2pt}
  \setlength{\dblfloatsep}{8pt plus 2pt minus 2pt}
  \setlength{\intextsep}{8pt plus 2pt minus 2pt}
  \captionsetup{skip=4pt}
  \setlength{\belowcaptionskip}{0pt}
}

\renewcommand{\TODO}[1]{}

\definecolor{cvprblue}{rgb}{0.21,0.49,0.74}
\usepackage[pagebackref,breaklinks,colorlinks,allcolors=cvprblue]{hyperref}

\title{AGLR: Attribute-Guided Latent Refinement for Fine-Grained Humanoid Motion Generation and Editing}

\author{Shuai Yuan$^{1}$ \quad Yuwei Gui$^{1,2}$ \quad Mingshuang Luo$^{2}$ \quad Zhuo Li$^{2}$\\
Siyuan Meng$^{3}$ \quad Wenhao Cui$^{3\dagger}$ \quad RuiBing Hou$^{2}$ \quad Bingpeng Ma$^{1\dagger}$\\[4pt]
$^{1}$University of the Chinese Academy of Sciences\\
$^{2}$Institute of Computing Technology, Chinese Academy of Sciences\\
$^{3}$Noetix Robotics \qquad $^{\dagger}$Corresponding authors\\[4pt]
\textbf{Project Page:} \href{https://syuan8889.github.io/AGLR-project}{\texttt{https://syuan8889.github.io/AGLR-project}}
}

\begin{document}
\maketitle
\begin{strip}
\centering
\includegraphics[width=\linewidth]{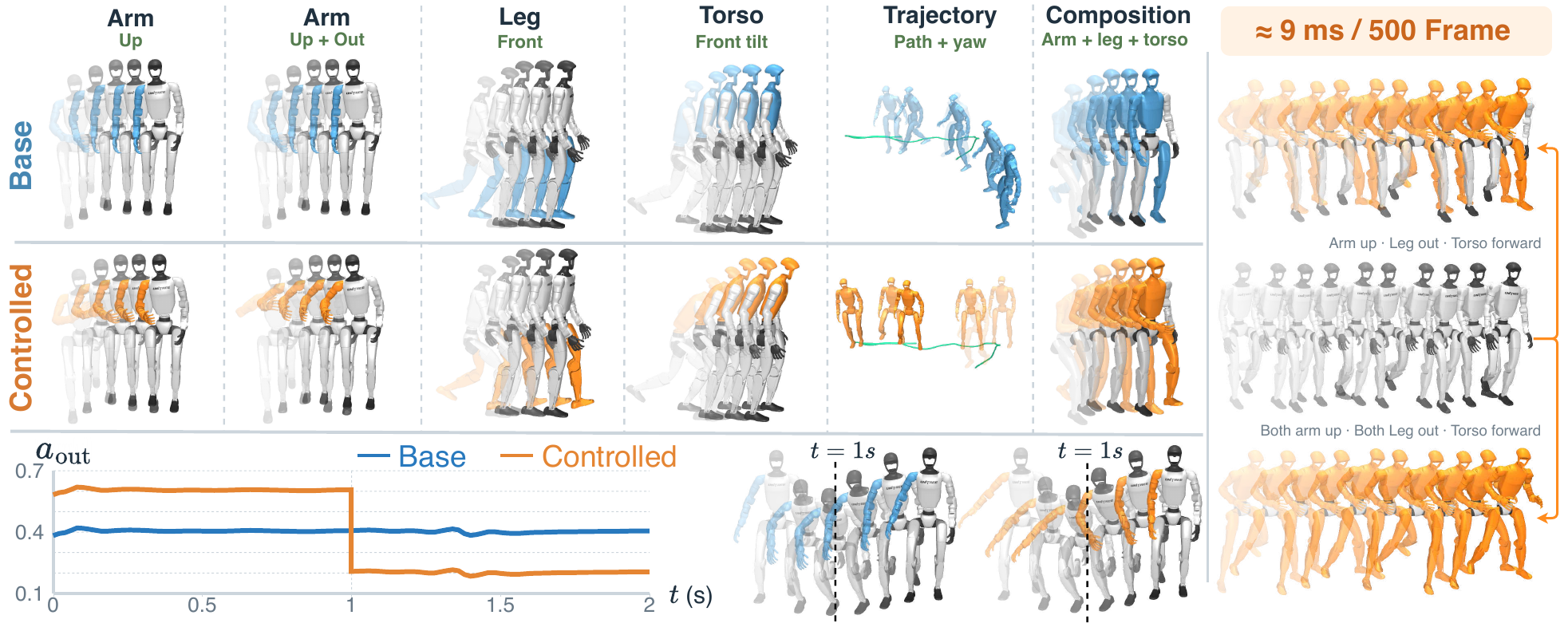}
\captionsetup{hypcap=false}
\captionof{figure}{
\textbf{Fine-grained, fast, composable, and temporally adaptive humanoid motion generation.} Our method enables intuitive control over individual limbs along three spatial axes, torso orientation, and root trajectories. These controls can be freely combined and varied over time, enabling diverse, precise, and localized motion refinements within a single motion instance.
}
\captionsetup{hypcap=true}
\label{fig:teaser}%
\end{strip}

\begin{abstract}
Humanoid motion generation must accommodate evolving task requirements, where localized adjustments should satisfy new geometric constraints without disrupting desired motion details. However, language-based control lacks geometric precision, while joint-level constraints require explicit spatial targets. Local geometric constraints are also underdetermined, complicating target satisfaction and source preservation. We present AGLR (Attribute-Guided Latent Refinement), a unified framework for fine-grained humanoid motion generation and editing through source-anchored refinement. A compact geometric interface supports composable, time-varying control of limb extents, torso orientation, and root trajectories. Given a generated or encoded base latent, an Attribute Residual Adapter predicts a single-pass correction, with dual-level supervision enforcing target matching and selective preservation across attributes and motion coordinates. On HumanML3D-G1, AGLR achieves the highest controllable-generation and preservation-constrained editing success rates among evaluated methods. Editing take only 9.1 ms for a cached 500-frame motion, without additional sampling or test-time optimization. Zero-shot evaluation on BONES-SEED-G1 further demonstrates transfer without fine-tuning.
\end{abstract}

\section{Introduction}
\label{sec:intro}

Humanoid robots operating in dynamic environments must adapt their motions to changing task requirements. Recent methods generate diverse whole-body motions from natural language instructions~\cite{robotmdm2024, schur2025, textop2026, echo2026, omg2026, safeflow2026}, 
providing references for motion tracking policies. Many tasks, however, require only localized modifications to an existing reference. 
For example, a humanoid may need to raise one arm while walking or adjust its torso orientation while maintaining its natural gait. Such tasks call for fine-grained motion editing that modifies selected geometric properties while preserving the rest of the reference motion.

The first challenge lies in specifying geometric edits with both precision and compositionality. While semantic control can describe high-level actions for individual body parts~\cite{finemogen2023, stmc2024}, it offers limited direct control over the magnitude and timing of a geometric change, such as precisely how far an arm should extend forward at each phase of a step. Conversely, kinematic control specifies spatial targets through joint positions~\cite{omnicontrol2024, sfcontrol2025, maskcontrol2025}, full-body poses~\cite{priormdm2024, condmdi2024}, or task-space trajectories~\cite{gmd2023, tlcontrol2024, kimodo2026}. Depending on the edit, these inputs often demand burdensome kinematic details beyond what the task naturally specifies. This gap motivates a part-level geometric interface that supports numerical, time-varying targets and enables seamless composition of edits across different body parts.

The second challenge is to satisfy these targets efficiently while faithfully preserving the details of the source motion. Multiple whole-body trajectories can satisfy the same set of geometric targets; for instance, prescribing the forward extent of one arm during walking leaves the gait and the movement of the opposite arm underdetermined. Existing source-conditioned editors~\cite{motionfix2024, simmotionedit2025, motionlab2025} leverage a reference trajectory to guide the transformation, where unselected attributes and motion coordinates are preserved. 
However, diffusion-based editors~\cite{omnime2026, motionfix2024, trace2026} often require iterative sampling, whereas optimization-based methods~\cite{dno2024, rgdno2026} rely on repeated gradient updates, significantly increasing inference latency. These limitations highlight the need for a framework that directly predicts motion corrections in a single forward pass, while providing robust supervision for both target satisfaction and source preservation.

In this work, we introduce \textbf{Attribute-Guided Latent Refinement (AGLR)}, a framework for fine-grained 
geometric adaptation of generated or existing humanoid robot reference motions.
AGLR specifies edits through the \textbf{Whole-body Geometric Attribute Representation (WGAR)}, a 21-dimensional per-frame descriptor capturing directional limb extents, torso orientation, and root motion. A channel-wise mask dictates which attributes to edit, allowing time-varying targets to be combined across body parts. To preserve source motion details left underdetermined by the selected targets, AGLR refines the source trajectory within a robot-native latent space learned from dynamics-grounded motions.
An \textbf{Attribute Residual Adapter (ARA)} predicts an additive correction to the source latent in a single forward pass. The training objective encourages the edited motion to satisfy the selected targets while preserving unselected attributes and motion coordinates. Finally, a downstream motion tracking policy can use the refined trajectory as a reference to execute the motion on a physical humanoid robot.

We evaluate AGLR on HumanML3D-G1 for controllable generation and preservation-constrained editing. AGLR achieves state-of-the-art success rates among compared methods, reaching 80.34\% for controllable generation and 76.81\% and 77.28\% for editing under attribute and joint preservation constraints, respectively. Furthermore, latent refinement and decoding take merely 9.1 ms for a 500-frame motion. Zero-shot evaluations on BONES-SEED-G1 demonstrate robust cross-dataset transfer without fine-tuning. Finally, real-world experiments on a Unitree G1 humanoid validate that AGLR can selectively and reliably modify whole-body motions during physical execution.

\section{Related Work}
\label{sec:related_work}


\noindent\textbf{Human and Humanoid Motion Generation.}
Existing human text-to-motion generation typically relies on variational models~\cite{temos2022}, diffusion processes in motion or latent spaces~\cite{mdm2023,motiondiffuse2024,flame2023,mld2023}, and discrete motion tokens~\cite{t2mgpt2023,motiongpt2023,momask2024}.
Extending this paradigm to humanoid robots requires representing motions in the robot’s configuration space. HARMON~\cite{harmon2024} transfers generated human motions via  kinematic retargeting, whereas these works~\cite{robotmdm2024,echo2026,textop2026} directly generate robot-space motion references.
Humanoid-X~\cite{humanoidx2024} and SCHUR~\cite{schur2025} use video-derived motions to expand training data, while OMG~\cite{omg2026} broadens the conditioning interface with multimodal inputs.
Building upon robot-space motion generation, AGLR employs a text-conditioned latent flow model to generate a base motion for subsequent geometric refinement.


\noindent\textbf{Fine-Grained Motion Control.}
Prior work enables body-part control through part-aware representations~\cite{partigen2026}, simultaneous part-level actions~\cite{sinc2023,frankenmotion2026}, spatial and temporal composition~\cite{finemogen2023,stmc2024}, and part-specific instructions~\cite{motionctrl2025}.
Language describes the intended actions, while spatial control interfaces provide numerical targets for trajectories~\cite{gmd2023,tlcontrol2024}, poses or motion regions~\cite{priormdm2024,condmdi2024}, and selected joint positions~\cite{omnicontrol2024,sfcontrol2025,maskcontrol2025}. Some works~\cite{kimodo2026,ardy2026} combine language with pose and path constraints.
These interfaces offer precise control over specified locations, but controlling a limb’s overall directional extent still requires translating that goal into point-level targets.
Our geometric interface instead exposes directional limb extents, torso orientation, and root motion as compact, directly controllable attributes. Attributes can be selected, combined across body parts, and varied over time without specifying complete joint trajectories.

\noindent\textbf{Motion Editing.}
Motion editing adapts a source motion to new spatial or semantic constraints while preserving its characteristics. 
MotionFix~\cite{motionfix2024} and SimMotionEdit~\cite{simmotionedit2025} learn instruction-guided edits, MotionLab~\cite{motionlab2025} unifies motion generation and editing, and SALAD~\cite{salad2025} uses attention for fine-grained semantic control.
TRACE~\cite{trace2026} preserves source coordinates through reference projection, while  UniMoFlow~\cite{unimoflow2026} refines source latents using conditioned velocity differences.
Other methods~\cite{tlcontrol2024,dno2024,rgdno2026} iteratively optimize motion latents or diffusion noise via repeated gradient updates, which add inference cost. 
MotionCLIP~\cite{motionclip2022} and MoDi~\cite{modi2023} support direct latent editing, but their semantic controls offer limited support for precise geometric targets.


\begin{figure*}[!t]
\centering
\includegraphics[width=\linewidth]{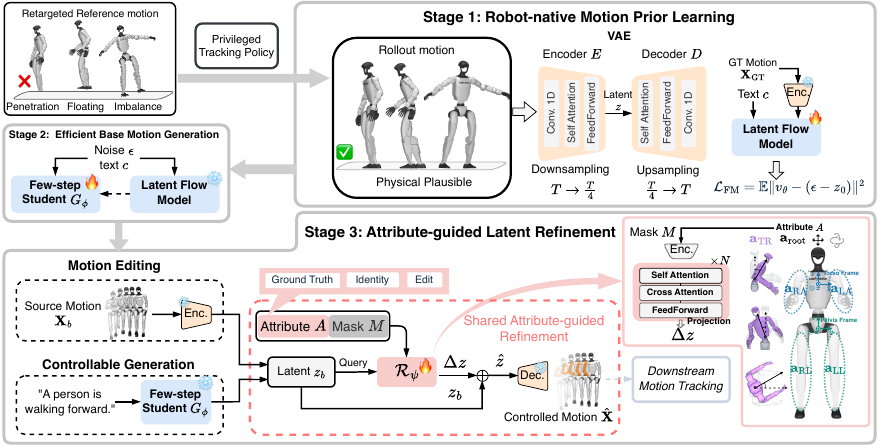}
\caption{\textbf{AGLR framework overview.}
AGLR learns a robot-native motion prior from policy rollouts  and distills it into a few-step generator. The Attribute Residual Adapter ($\mathcal R_\psi$) refines generated or source motion latents in a single pass to meet selected geometric targets while preserving unedited motion details.
}
\label{fig:framework}
\vspace{-5pt}
\end{figure*}
\section{Method}
\label{sec:method}

\textbf{Problem Formulation.}
We investigate fine-grained geometric adaptation of humanoid whole-body motions.
Given a base motion $\mathbf{X_b}$, target attribute sequence $\mathbf{A}\in\mathbb R^{T\times K}$, and a binary channel mask $\mathbf{M}\in\{0,1\}^{K}$, our goal is to generate an adapted motion $\mathbf{\hat X}$ that satisfies the selected target attributes while preserving unselected attributes and other motion details. Here, $T$ counts valid frames and $K$ denotes the number of attribute channels. Formally, the adaptation process is expressed as $\mathbf{\hat X} = \mathcal{H}(\mathbf{X_b}, \mathbf{A}, \mathbf{M})$, where $\mathcal{H}$ denotes the geometric adaptation function.


\noindent\textbf{Overview.}
To achieve this task, we propose AGLR, a unified framework for fine-grained geometric adaptation of humanoid whole-body motions. As illustrated in Fig.~\ref{fig:framework}, AGLR operates across three stages.

\noindent
\textbf{Stage 1: Robot-native Motion Prior Learning.} We first train a robot-native variational autoencoder (VAE) on physically plausible motions to learn a compact latent representation tailored to the robot's kinematic structure. Building upon this representation, we further learn a latent flow prior to capture the underlying distribution of robot motions.

\noindent
\textbf{Stage 2: Efficient Base Motion Generation.} To reduce the cost of base motion generation, we distill the pretrained latent flow model into a few-step generator, enabling efficient sampling while retaining the learned motion prior.

\noindent
\textbf{Stage 3: Attribute-guided Latent Refinement.}
To facilitate fine-grained geometric adaptation, we introduce two components: \textit{Whole-body Geometric Attribute Representation (WGAR)} and \textit{Attribute Residual Adapter (ARA)}. Specifically, WGAR provides an interpretable and compositional interface for specifying geometric targets across different body parts, while ARA performs attribute-conditioned refinement in the robot-native latent space. By introducing target-specific geometric adjustments in the latent space, ARA aims to satisfy the specified attributes while preserving non-target motion details.

During inference, AGLR first obtains an initial motion latent either from the distilled generator or by encoding an existing motion with the pretrained VAE encoder.
Given \(\left(\mathbf{A},\mathbf{M}\right)\), ARA then refines the initial latent to produce an attribute-consistent representation, which is subsequently decoded by the VAE to obtain the adapted motion \(\mathbf{\hat{X}}\).

\subsection{Whole-body Geometric Attributes}
\label{sec:3.1}
\textbf{Robot Motion Representation.}
We represent humanoid robot motions using trajectories generated by a privileged tracking policy in simulation~\cite{omnitrack2026}. 
Similar to ECHO~\cite{echo2026}, we adopt RPR39 as per-frame motion representation, denoted as $\mathbf{x}_t \in \mathbb{R}^{39}$. It consists of 29 joint angles, two heading-local horizontal root displacement increments, one root height, one yaw increment, and a 6D heading-local root rotation representation. 
Given this representation, we reconstruct the root trajectory from the displacement and yaw increments, and subsequently apply forward kinematics to recover the 3D positions $\mathbf p_{t,j}\in\mathbb R^3$ and orientations $\mathbf R_{t,j}\in SO(3)$ of each body element $j$ at frame $t$.

\noindent
\textbf{Limb Attributes.} 
We describe each arm and leg using three directional attributes: Front, Out, and Up, defined in the torso or pelvis coordinate frame. For limb \(k\), let \(\mathbf p_{t,a_k}\), \(\mathbf p_{t,m_k}\), and \(\mathbf p_{t,e_k}\) denote the positions of its anchor, middle joint, and endpoint. We use the average position of the middle joint and endpoint to characterize the overall limb configuration:
\begin{equation}
\mathbf a_{t,k}
=
\frac{1}{\ell_k}
\mathbf S_k\mathbf Q_{t,k}^{\top}
\left(
\frac{\mathbf p_{t,m_k}+\mathbf p_{t,e_k}}{2}
-\mathbf p_{t,a_k}
\right)
\end{equation}
Here, \(\mathbf Q_{t,k}\) denotes the neutral-aligned torso orientation for arms or pelvis orientation for legs, and \(\ell_k\) is the limb-length normalization factor. The local axes correspond to forward, outward, and upward directions. To unify lateral semantics across both sides, we set \(\mathbf S_k=\operatorname{diag}(1,1,1)\) for left limbs and \(\operatorname{diag}(1,-1,1)\) for right limbs, ensuring that positive lateral values consistently indicate outward extension.

\noindent\textbf{Torso Attributes.}
We represent pelvis-relative torso orientation using three attributes: \textit{Forward Tilt}, \textit{Rightward Tilt}, and \textit{Heading}. Given the global pelvis and torso orientations, we compute the relative rotation as
 $\mathbf R_{t,\mathrm{rel}}=\mathbf R_{t,\mathrm{pelvis}}^\top\mathbf R_{t,\mathrm{torso}}$, with up and forward axes $\mathbf u_t=\mathbf R_{t,\mathrm{rel}}\mathbf e_z$ and $\mathbf v_t=\mathbf R_{t,\mathrm{rel}}\mathbf e_x$. Here, $\mathbf e_z$ and $\mathbf e_x$ are unit coordinate vectors. Using $\alpha(y,x)=\operatorname{atan2}(y,x)$, the torso attributes are defined as:
\begin{equation}
\scalebox{0.95}{$\displaystyle
\mathbf a_{t,\mathrm{torso}} =
[\alpha(u_{t,x},u_{t,z}),
 \alpha(-u_{t,y},u_{t,z}),
 \alpha(v_{t,y},v_{t,x})]^\top
$}
\end{equation}
The three components represent forward tilt, rightward tilt, and heading relative to the pelvis, respectively, making them invariant to global heading changes.

\noindent
\textbf{Root Attributes.} 
We represent root motion relative to its initial heading. Let \(\mathbf p_t\) and \(\mathbf R_t\) denote the global root position and orientation, and \(\mathbf p_0\) and \(\mathbf H_0\) the initial position and yaw rotation. The initial-heading coordinates are given by
$\tilde{\mathbf p}_t=\mathbf H_0^\top(\mathbf p_t-\mathbf p_0)$ and $\tilde{\mathbf R}_t=\mathbf H_0^\top\mathbf R_t$. We therefore define
\begin{equation}
\mathbf a_{t,\mathrm{root}}=[\tilde p_{t,x}, \tilde p_{t,y}, p_{t,z}, \operatorname{rpy}(\tilde{\mathbf R}_t)^\top]^\top
\end{equation}
where \(\operatorname{rpy}(\cdot)\) returns the roll, pitch, and yaw angles. This captures horizontal displacement and orientation in the initial-heading frame while preserving absolute root height.

To obtain a unified representation, we concatenate the limb, torso, and root attributes at each frame:
$\mathbf a_t =
\left[
\mathbf a_{t,1}^{\top},\ldots,
\mathbf a_{t,K_l}^{\top},
\mathbf a_{t,\mathrm{torso}}^{\top},
\mathbf a_{t,\mathrm{root}}^{\top}
\right]^{\top}$,
where \(K_l\) denotes the number of limb attribute groups. Stacking the per-frame vectors yields the attribute sequence \(\mathbf{A}=[\mathbf a_1,\ldots,\mathbf a_T]^\top\in\mathbb R^{T\times K}\), \(K\) denotes the total number of attribute channels.

\subsection{Attribute Residual Adapter}
\noindent
\textbf{Latent Residual Refinement.}
The required geometric adjustments depend on the underlying whole-body configuration. To account for this dependency, we introduce the Attribute Residual Adapter (ARA) to perform attribute-conditioned latent refinement in the shared robot-native latent space.
Let $\mathbf{z_b}$ denote the base latent, either produced by the generator or obtained by encoding a source motion. ARA takes \(\mathbf{z_b}\), the target attributes \(\mathbf{A}\), and the selection mask \(\mathbf{M}\) as inputs to predict a latent residual:
\begin{equation}
\scalebox{0.9}{$\displaystyle
\begin{aligned}
\mathbf{\hat X}&=D\left(\mathbf{z_b}+\mathcal R_\psi(\mathbf{z_b},\mathbf{A},\mathbf{M})\right),\\
\mathcal R_\psi(\mathbf{z_b},\mathbf{A},\mathbf{M})
&=W_{\mathrm{out}}\left(
\mathcal T\left(P(\mathbf{z_b}),\Phi(\mathbf{M}\odot \mathbf{A},\mathbf{M})\right)\right).
\end{aligned}
$}
\label{eq0}
\end{equation}
Here, $\mathcal{R}_\psi$ represents the ARA parameterized by $\psi$. \(P\) projects the base latent into motion tokens, while \(\Phi\) encodes the masked target attributes and selection mask into conditioning tokens. The Transformer \(\mathcal T\) employs self-attention to capture motion context and cross-attention to incorporate attribute conditions. Finally, \(W_{\mathrm{out}}\) maps the output features to the latent residual, which is added to the base latent and decoded by the pretrained VAE decoder \(D\). Rather than regenerating the entire motion, this residual refinement enables targeted geometric adaptation while preserving the original motion context and non-target details.

\noindent
\textbf{Attribute-level Supervision.}
We enforce geometric consistency directly in the attribute space, supervising target satisfaction and preservation.
Specifically, given the refined attributes \(\mathbf{\hat A}=\operatorname{WGAR}(\mathbf{\hat X})\), base attributes \(\mathbf{A_b}=\operatorname{WGAR}(\mathbf{X_b})\), target attribute sequence \(\mathbf{A}\), and selection mask \(\mathbf{M}\), we encourage selected attributes to match their targets while preserving unselected attributes relative to the base motion. The resulting loss is defined as:
\begin{equation}
\resizebox{0.90\columnwidth}{!}{$\displaystyle
\mathcal L_{\mathrm{attr}}\left(\mathbf{\hat X},\mathbf{X_b},\mathbf{A},\mathbf{M}\right)
=\langle e(\mathbf{\hat A},\mathbf{A})\rangle_{\mathcal S}
+\langle e(\mathbf{\hat A},\mathbf{A_b})\rangle_{\bar{\mathcal S}},
$}
\label{eq5}
\end{equation}
where \(\mathcal S\) and \(\bar{\mathcal S}\) denote the selected and unselected attribute channels indicated by \(\mathbf{M}\), respectively. The operator \(\langle\cdot\rangle\) independently averages each term over valid frames. 
The channel-wise error \(e\) is tailored to different attribute types, employing normalized positional errors, periodic angular errors, and rotation-aware errors for root orientation. Details are provided in the supplement.

\noindent
\textbf{Motion-level Supervision.}
Although attribute-level constraints preserve the geometric properties encapsulated by WGAR, they are insufficient to fully govern joint configurations and motion details. We therefore introduce an additional preservation constraint directly on the original motion representation, formulated as
\begin{equation}
\mathcal L_{\mathrm{coord}}(\mathbf{\hat X},\mathbf{X_b},\mathbf{M})
=\frac{1}{T|\mathcal J|}
\sum_{t=1}^{T}\sum_{j\in\mathcal J}
(\hat x_{t,j}-x_{b,t,j})^2,
\label{eq6}
\end{equation}
where $\hat x$ and $x_b$ denote the refined and base motion coordinates respectively, and $\mathcal J=\mathcal J_{\mathrm{prot}}(M)$ is the protected coordinate set.
The protected coordinate set is determined by the selected geometric attributes, allowing coordinates associated with targeted modifications to change while constraining the remaining to preserve the original motion.

\subsection{Training Strategy}
To achieve fine-grained geometric adaptation while preserving non-target motion details, we train ARA using three complementary objectives: paired-target supervision, identity supervision, and selective editing. These objectives respectively enable target adaptation, identity preservation, and selective geometric modification.

During training, $\mathbf X_b=D(\mathbf z_b)$ denotes the
decoded base used for preservation supervision.
For source-motion editing, $\mathbf X_b$
denotes the original input motion, encoded as
$\mathbf z_b=E(\mathbf X_b)$.
For each task \(\tau \in \{\mathrm{GT},\mathrm{ID},\mathrm{Edit}\}\), ARA takes \((\mathbf{z_b},\mathbf{A_\tau},\mathbf{M_\tau})\) as input, where \(\mathbf{A_\tau}\) and \(\mathbf{M_\tau}\) denote the target attribute sequence and selection mask, respectively. The refined latent and motion are obtained as:
\begin{equation}
\begin{aligned}
\mathbf{\hat z_\tau}&=\mathbf{z_b}+\mathcal R_\psi(\mathbf{z_b},\mathbf{A_\tau},\mathbf{M_\tau}), \\
\mathbf{\hat X_\tau}&=D(\mathbf{\hat z_\tau}),
\ \tau \in \{\mathrm{GT},\mathrm{ID},\mathrm{Edit}\}.
\end{aligned}
\end{equation}
We define the latent reconstruction loss as
$\mathcal L_z(\mathbf{\hat z},\mathbf{z})=\|\mathbf{\hat z}-\mathbf{z}\|^2$.  
To discourage physically implausible joint configurations, we apply a joint-limit penalty to all tasks:
\begin{equation}
\mathcal L_q(\mathbf{\hat X})=\frac{1}{T}\sum_{t,j}
\left([\hat q_{t,j}-q_j^{\max}]_+^2+
[q_j^{\min}-\hat q_{t,j}]_+^2\right),
\label{eq8}
\end{equation}
where \(\hat q_{t,j}\) denotes the predicted joint angle, \(q_j^{\min}\) and \(q_j^{\max}\) are its corresponding limits, and \([u]_+=\max(u,0)\).

\noindent
\textbf{Paired-target and Identity Supervision.}
We first introduce two complementary supervision settings that share the same loss formulation but differ in target construction: i)
For paired-target supervision, we construct training pairs consisting of a base motion  and its corresponding ground-truth target \(\mathbf{X_{\mathrm{GT}}}\). We extract the target attributes as \(\mathbf{A_{\mathrm{GT}}}=\operatorname{WGAR}(\mathbf{X_{\mathrm{GT}}})\) and set \(\mathbf{M_{\mathrm{GT}}}=\mathbf 1\) to supervise all attribute channels. The target motion is encoded as \(\mathbf{z_{\mathrm{GT}}}=E(\mathbf{X_{\mathrm{GT}}})\) for latent-level supervision.
ii) For identity supervision, we replace the ground-truth target with the base motion, setting \(\mathbf{A_{\mathrm{ID}}}=\mathbf{A_b}\) and \(\mathbf{z_{\mathrm{ID}}}=\mathbf{z_b}\), while randomly sampling an attribute mask \(\mathbf{M_{\mathrm{ID}}}\). This encourages ARA to preserve the original motion under different attribute-selection patterns.
Both settings are optimized by enforcing latent reconstruction, geometric attribute consistency, and  joint-limit compliance:
\begin{equation}
\begin{aligned}
\mathcal L_\tau
={}&\mathcal L_z(\mathbf{\hat z_\tau},\mathbf{z_\tau})
+\mathcal L_{\mathrm{attr}}(
\mathbf{\hat X_\tau},\mathbf{X_b},\mathbf{A_\tau},\mathbf{M_\tau})\\
&+\lambda_q\mathcal L_q(\mathbf{\hat X_\tau}),
\quad \tau \in\{\mathrm{GT},\mathrm{ID}\},
\end{aligned}
\end{equation}

\noindent
\textbf{Selective Editing.}
To enable fine-grained geometric adaptation, we construct synthetic targets \(\mathbf{A_{\mathrm{Edit}}}\) by modifying selected attributes of the base motion while preserving unselected ones induced by \(\mathbf{M_{\mathrm{Edit}}}\) . Unlike the previous two tasks, no paired edited motion or corresponding target latent is available, so we omit latent reconstruction and directly supervise geometric adaptation and motion preservation:
\begin{equation}
\begin{aligned}
\mathcal L_{\mathrm{Edit}}
={}&\mathcal L_{\mathrm{attr}}(\mathbf{\hat X_{\mathrm{Edit}}},\mathbf{X_b},\mathbf{A_{\mathrm{Edit}}},\mathbf{M_{\mathrm{Edit}}}) \\
&+\mathcal L_{\mathrm{coord}}(\mathbf{\hat X_{\mathrm{Edit}}},\mathbf{X_b},\mathbf{M_{\mathrm{Edit}}})
+\lambda_q\mathcal L_q(\mathbf{\hat X_{\mathrm{Edit}}}).
\end{aligned}
\end{equation}
Here, attribute-level supervision enforces the target geometry, while coordinate-level supervision preserves non-target motion details. 

We combine the three training objectives using a weighted sum: $
\mathcal L=\lambda_{\mathrm{GT}}\mathcal L_{\mathrm{GT}}
+\lambda_{\mathrm{ID}}\mathcal L_{\mathrm{ID}}
+\lambda_{\mathrm{Edit}}\mathcal L_{\mathrm{Edit}}$.
During training, only ARA parameters \(\psi\) are optimized, while the VAE, distilled generator, and WGAR mapping remain frozen.

\subsection{Efficient Generation and Deployment}
\noindent
\textbf{Latent Flow Prior and Distillation.}
We employ a text-conditioned latent flow model to generate base motion latents. To accelerate inference, we distill the pretrained flow model into a four-step student using DMD2~\cite{yin2024dmd2}. The distilled generator is frozen and provides base motion proposals for ARA, ensuring consistency between training and inference distributions. Details of the VAE, latent flow model, and distillation procedure are provided in the supplement.

\noindent\textbf{Latent Refinement and Cached Editing.}
Given a base latent \(\mathbf{z_b}\), obtained either from the distilled generator or by encoding an existing motion, ARA performs attribute-guided refinement according to the target attributes \(\mathbf{A}\) and selection mask \(\mathbf{M}\) (Eq.~\ref{eq0}). This process requires only one ARA forward pass and one VAE decoding pass, without additional flow sampling. Moreover, the base latent can be cached and reused for successive edits.

\noindent\textbf{Robot Deployment and Motion Tracking.}
For deployment, given a source motion \(\mathbf{X_b}\), we obtain its base latent through the pretrained VAE encoder, i.e., \(\mathbf{z_b}=E(\mathbf{X_b})\), and refine it using the same attribute-guided process. The resulting motion \(\mathbf{\hat X}\) serves as a reference for a downstream tracking policy, which generates control actions to execute the edited motion on the robot. This decouples geometric adaptation from closed-loop motion execution, allowing the same tracking policy to handle different geometric edits.
\section{Experiments}
We evaluate AGLR across three core tasks: text-to-motion generation, fine-grained controllable generation, and source-conditioned motion editing. 
Complete training, evaluation, baseline adaptation details for controllable generation and motion editing are provided in the supplement, alongside extended editing comparisons, ablation studies, and additional hardware demonstrations.

\label{sec:exp}
\subsection{Experimental Setup}
We use HumanML3D-G1 as the primary benchmark. To assess cross-dataset generalization, we evaluate models trained on HumanML3D-G1 in zero-shot on BONES-SEED-G1.

\noindent\textbf{HumanML3D-G1.} 
We construct HumanML3D-G1 by retargeting HumanML3D~\cite{humanml3d2022} motions to Unitree G1 and tracking them in simulation with a
pretrained privileged policy~\cite{omnitrack2026} to mitigate
retargeting artifacts. After filtering rollouts with large tracking errors, we preserve the original splits, retaining 19,728 training and 3,766 test samples. Motions use RPR39 at
50 Hz, with normalization statistics computed using the training split.

\noindent\textbf{BONES-SEED-G1.} We use the official G1-retargeted motions from BONES-SEED~\cite{bonesseed} and track them in simulation using the same pretrained privileged policy as for HumanML3D-G1. For zero-shot evaluation, we randomly sample 1,000 resulting rollouts from each of the content and repetition test splits, yielding 2,000 evaluation motions.


\begin{table}[t]
\centering
\captionsetup{font=footnotesize,skip=3pt}
\renewcommand{\arraystretch}{0.95}
\setlength{\tabcolsep}{1.5pt}
\fontsize{7}{8}\selectfont
\resizebox{\columnwidth}{!}{%
\begin{tabular}{@{}llrrrrrrr@{}}
\toprule
Method & Ref.
& FID$\downarrow$
& R@1$\uparrow$
& R@2$\uparrow$
& R@3$\uparrow$
& MD$\downarrow$
& Div.
& Lat.$\downarrow$\\
\midrule
TextOp~\cite{textop2026} & arXiv'26
& 0.1267 & 42.95 & 59.96 & 69.85
& 1.1506 & 1.3155 & 1592.4 \\
ECHO~\cite{echo2026} & arXiv'26
& 0.0437 & 54.91 & 72.38 & 79.67
& 1.1131 & 1.3294 & 229.4 \\
MDM~\cite{mdm2023} & ICLR'23
& 0.1068 & 46.93 & 64.16 & 73.10
& 1.1404 & 1.3199 & 315.5 \\
MoMask~\cite{momask2024} & CVPR'24
& 0.0448 & 49.01 & 65.06 & 72.25
& 1.1294 & 1.3425 & \underline{180.4} \\
MotionHiFlow~\cite{motionhiflow2026} & CVPR'26
& 0.0590 & 58.46 & 74.32 & 82.14
& \underline{1.0865} & 1.3284 & 297.4 \\
\midrule
AGLR (50 steps) & ---
& \textbf{0.0154} & \textbf{62.07}
& \textbf{77.54} & \textbf{84.54}
& \textbf{1.0752} & 1.3222 & 1206.3 \\
AGLR (4 steps) & ---
& \underline{0.0184} & \underline{59.96}
& \underline{75.69} & \underline{83.07}
& \underline{1.0865} & 1.3222
& \textbf{55.2} \\
\bottomrule
\end{tabular}%
}

\caption{\textbf{Text-to-humanoid motion generation.}
Text-only evaluation on the HumanML3D-G1 test split.
R@K is reported in $\%$ and latency in $\mathrm{ms}$.}
\label{tab:generation}
\vspace{-10pt}
\end{table}

\subsection{Text-to-Motion Generation}

\noindent\textbf{Baselines.}
We compare AGLR with G1-native methods (ECHO~\cite{echo2026} and TextOp~\cite{textop2026}) and methods adapted to G1 (MDM~\cite{mdm2023}, MoMask~\cite{momask2024}, and MotionHiFlow~\cite{motionhiflow2026}). All methods use the same motion representation, dataset split, and evaluation protocol. To evaluate our generator in isolation, we use text as the sole input to both the 50-step teacher and the four-step student and bypass ARA in both cases.

\noindent\textbf{Metrics.}
We report R-Precision (R@1, R@2, and R@3) with 32 retrieval candidates, Fréchet Inception Distance (FID), Matching Distance (MD), and Diversity (Div.). We compute all four metrics using the same frozen G1-specific TMR~\cite{tmr2023} evaluator. End-to-end inference latency (Lat.) is measured in milliseconds and includes text encoding, sampling, decoding, and pose conversion. 

\noindent\textbf{Quantitative comparisons.}
In Table~\ref{tab:generation}, the 50-step teacher achieves the lowest FID and MD and the highest R-Precision at all three cutoffs. Distillation reduces end-to-end latency from 1,206.3 to 55.2 ms, a \(21.85\times\) speedup. The four-step student outperforms all external baselines on FID and R-Precision and matches the best baseline (MotionHiFlow~\cite{motionhiflow2026}) on MD metric. We use the four-step student to generate base motions for subsequent experiments.

\begin{figure*}[t]
    \centering
    \includegraphics[width=\linewidth]{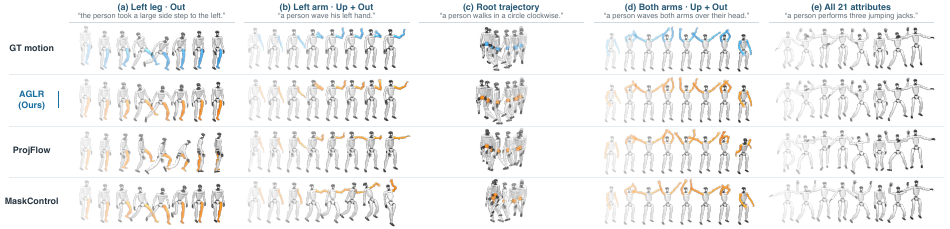}
    \caption{\textbf{Fine-grained controllable generation.} Columns show five distinct control requests specified above the motions, and rows compare Ground Truth (GT), AGLR, ProjFlow~\cite{projflow2026}, and MaskControl~\cite{maskcontrol2025} using identical text prompts and ground-truth-derived targets. Highlighted regions denote the actively controlled body parts.}
    \label{fig:qualitative_comparison}
\end{figure*}

\subsection{Fine-Grained Controllable Generation}
\noindent\textbf{Task.}
We evaluate on the HumanML3D-G1 test split using $\mathbf{A}_{\mathrm{GT}}=\operatorname{WGAR}(\mathbf{X}_{\mathrm{GT}})$, with $\mathbf{M}$ selecting a single attribute, multiple
attributes within or across body parts, or all channels.  $(\mathbf{A}_{\mathrm{GT}},\mathbf{M})$ guide ARA and condition the baselines. All methods share text prompts and motion lengths.

\noindent\textbf{Baselines and adaptation.}
We compare with six representative controllable motion generation baselines, OmniControl~\cite{omnicontrol2024}, MotionLCM~\cite{motionlcm2024}, MaskControl~\cite{maskcontrol2025}, TLControl~\cite{tlcontrol2024}, SFControl~\cite{sfcontrol2025}, and ProjFlow~\cite{projflow2026}, on HumanML3D-G1 using the RPR39 representation.
We retrain each baseline on HumanML3D-G1 using RPR39 and adapt it to accept the same 
$(\mathbf{A}_{\mathrm{GT}},\mathbf{M})$
as AGLR.

\begin{table}[h]
\centering
\captionsetup{font=footnotesize,skip=3pt}
\renewcommand{\arraystretch}{0.95}
\setlength{\tabcolsep}{1.2pt}
\fontsize{7}{8}\selectfont
\resizebox{\columnwidth}{!}{%
\begin{tabular}{@{}llrrrrrrrrr@{}}
\toprule
Method & Ref.
& TAE$\downarrow$ & ASR$\uparrow$
& FID$\downarrow$
& R@1$\uparrow$ & R@2$\uparrow$ & R@3$\uparrow$
& MD$\downarrow$ & Div. & Lat.$\downarrow$ \\
\midrule
OmniControl~\cite{omnicontrol2024} & ICLR'24
& 0.8695 & 17.20 & 0.1258
& 42.74 & 58.92 & 67.95
& 1.1600 & 1.3150 & 3132.79 \\
TLControl~\cite{tlcontrol2024} & ECCV'24
& 0.5871 & 29.79 & 0.1431
& 25.11 & 34.13 & 39.69
& 1.2552 & 1.3085 & 723.87 \\
MotionLCM~\cite{motionlcm2024} & ECCV'24
& 0.3629 & 40.17 & 0.0773
& 44.15 & 60.18 & 68.72
& 1.1253 & 1.2783 & \textbf{85.95} \\
MaskControl~\cite{maskcontrol2025} & ICCV'25
& 0.6374 & 25.66 & \underline{0.0303}
& \underline{49.95} & \underline{66.05}
& \underline{74.52}
& \underline{1.1237} & 1.3343 & 4558.55 \\
SFControl~\cite{sfcontrol2025} & ICCV'25
& 0.3100 & 41.34 & 0.4884
& 21.79 & 31.14 & 37.53
& 1.2733 & 1.1501 & 1859.52 \\
ProjFlow~\cite{projflow2026} & CVPR'26
& \underline{0.1192} & \underline{73.66}
& 0.0776 & 45.94 & 61.81 & 69.87
& 1.1550 & 1.3369 & 13638.80 \\
\midrule
AGLR (ours) & ---
& \textbf{0.0740} & \textbf{80.34}
& \textbf{0.0225} & \textbf{56.04}
& \textbf{71.07} & \textbf{78.21}
& \textbf{1.1101} & 1.3244
& \underline{90.09} \\
\bottomrule
\end{tabular}%
}
\caption{\textbf{Fine-grained controllable generation.}
All methods share identical text prompts, motion lengths, masks, and ground-truth-derived attribute targets. ASR and R@K are reported in $\%$, and latency is reported in $\mathrm{ms}$.}
\label{tab:conditioned_generation}
\end{table}

\noindent\textbf{Metrics.}
Target Attribute Error (\textbf{TAE}) measures the root mean squared error over selected attribute channels after normalization using training-set statistics. Attribute Success Rate (\textbf{ASR}) calculates the percentage of frames in which all selected attributes simultaneously fall within their target tolerances: 0.05 for limb extents, 0.05 m for positions, and \(5^\circ\) for angles. We compute both metrics per sample and average them across the test set. We use the same TMR-based metrics as in text-to-motion generation to evaluate motion quality and measure end-to-end latency from input processing to controlled motion output.

\noindent\textbf{Quantitative comparisons.}
As shown in Table \ref{tab:conditioned_generation}, MotionLCM~\cite{motionlcm2024}  is slightly faster than AGLR (85.95 vs. 90.09 ms). MaskControl~\cite{maskcontrol2025} achieves a competitive FID. AGLR achieves the highest ASR and R-Precision and the lowest TAE, FID, and MD, combining more accurate attribute control with higher motion distribution fidelity and better text–motion alignment.

\noindent\textbf{Qualitative comparisons.}
Figure~\ref{fig:qualitative_comparison} showcases qualitative examples spanning individual limb control, cross-part composition, root trajectory control, and all-attribute conditioning. Across all settings, AGLR generates motions that remain consistent with the specified target attributes.
\begin{figure*}[t]
\centering
\includegraphics[width=\linewidth]{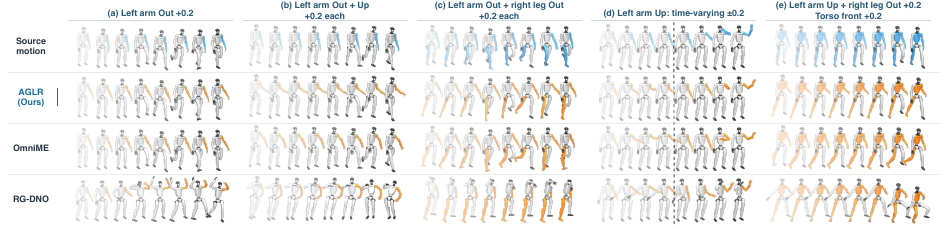}
\caption{\textbf{Selective editing.} Columns specify the requested attribute changes. Rows show the source and outputs of AGLR, OmniME~\cite{omnime2026}, and RG-DNO~\cite{rgdno2026}. Highlighted regions indicate selected parts. Limb offsets are normalized extents. Torso offsets are in radians.}
\label{fig:control_comparison}
\vspace{-5pt}
\end{figure*}

\subsection{Source-Conditioned Motion Editing}
\noindent\textbf{Task.}
All methods receive the same source motions $\mathbf{X_b}$, attribute mask, and target attribute values for each request.  We construct targets by randomly modifying selected attributes of $\operatorname{WGAR}(\mathbf{X_b})$ and use \(\mathbf{X_b}\) as the reference for measuring  preservation. Each test sample is assigned one fixed request from one of four categories: single-attribute, within-part, cross-part, and time-varying editing.

\noindent\textbf{Baselines and adaptation.}
We compare against five representative motion editing baselines:
TRACE~\cite{trace2026}, OmniME~\cite{omnime2026}, DNO~\cite{dno2024}, RG-DNO~\cite{rgdno2026}, and ScaleMoGen~\cite{scalemogen2026}. We adapt the baselines to RPR39 and the WGAR-based editing interface and retrain them on HumanML3D-G1. 

\noindent\textbf{Metrics.}
We use TAE to measure target adherence, and report Non-target Attribute Drift (\textbf{NAD}) and Non-target Joint Drift (\textbf{NJD}) for source preservation.
NAD measures normalized deviation from the source over unselected attributes, and NJD measures joint-space RMSE relative to the source over unedited body parts. We also report two preservation-constrained attribute success rates, \(\mathrm{ASR}_c^a\) and \(\mathrm{ASR}_c^j\):
\begin{equation}
\mathrm{ASR}_c^k(\tau)
=\frac{1}{N}\sum_{i=1}^{N}
\mathrm{ASR}_i\,\mathbf{1}[d_i^k\leq\tau],
\quad k\in\{a,j\},
\label{eq:asrc}
\end{equation}
where $\mathrm{ASR}_i$ is the frame-level success rate
for request $i$, and $d_i^a=\mathrm{NAD}_i$ and
$d_i^j=\mathrm{NJD}_i$ are the corresponding drift measures. 
A request contributes its ASR only when its drift falls within the threshold ($\tau$). We evaluate attribute and joint preservation independently across their respective threshold ranges.
Latency includes source processing, refinement, decoding,
and pose conversion.
\begin{table}[h]
\centering
\captionsetup{font=footnotesize,skip=3pt}
\renewcommand{\arraystretch}{0.95}
\setlength{\tabcolsep}{1.5pt}
\fontsize{7}{8}\selectfont
\resizebox{\columnwidth}{!}{%
\begin{tabular}{@{}llrrrrrr@{}}
\toprule
Method & Ref.
& TAE$\downarrow$
& NAD$\downarrow$
& NJD$\downarrow$
& ASR$_c^a\uparrow$
& ASR$_c^j\uparrow$
& Lat.$\downarrow$ \\
\midrule
TRACE~\cite{trace2026} & arXiv'26
& 0.2220 & \textbf{0.0557} & \textbf{0.000}
& \underline{63.80} & \underline{65.64} & 1009.22 \\
OmniME~\cite{omnime2026} & CVPR'26
& 0.2412 & 0.1226 & 1.383
& 61.51 & 64.32 & 969.72 \\
ScaleMoGen~\cite{scalemogen2026} & ECCV'26
& 0.8137 & 0.3867 & 7.989
& 10.05 & 6.58 & \underline{275.41} \\
DNO~\cite{dno2024} & CVPR'24
& 0.1514 & 0.5958 & 12.172
& 10.89 & 1.92 & 69721.37 \\
RG-DNO~\cite{rgdno2026} & CVPR'26
& \textbf{0.1008} & 1.2234 & 29.847
& 0.78 & 0.00 & 531482.50 \\
\midrule
AGLR (ours) & ---
& \underline{0.1431} & \underline{0.0584}
& \underline{0.681}
& \textbf{76.81} & \textbf{77.28}
& \textbf{13.15} \\
\bottomrule
\end{tabular}%
}

\caption{\textbf{Fine-grained source-conditioned motion editing.}
All methods share identical source motions and attribute requests.
ASR$_c^a$ and ASR$_c^j$ denote success rates under separate
NAD $\leq 0.30$ and NJD $\leq 3^\circ$ constraints, respectively, reported in $\%$, NJD in degrees,
and latency in $ms$.}

\label{tab:source_editing}
\vspace{-10pt}
\end{table}

\noindent\textbf{Quantitative comparisons.}
As shown in Table~\ref{tab:source_editing}, AGLR achieves the highest preservation-constrained success rates under both attribute and joint constraints (76.81\% and 77.28\%) and has the lowest latency.
TRACE~\cite{trace2026} uses reference projection to preserve protected joints exactly, yielding zero NJD and the lowest NAD. RG-DNO~\cite{rgdno2026} achieves lower TAE than AGLR but incurs substantially greater non-target drift. AGLR is approximately \(40{,}000\times\) faster than RG-DNO~\cite{rgdno2026} and \(5{,}300\times\) faster than DNO~\cite{dno2024} , highlighting the latency advantage of single-pass latent refinement over test-time optimization. 


\noindent\textbf{Refinement efficiency.}
With a cached source latent, ARA and decoding produce a
500-frame motion in 9.1 ms on an RTX 4090D at batch size one,
measured to GPU output. The 13.15 ms end-to-end latency in
Table~\ref{tab:source_editing} also includes source processing
and output conversion.

\noindent\textbf{Qualitative comparisons.}
Figure~\ref{fig:control_comparison} shows single-part, compositional, and time-varying edits. OmniME~\cite{omnime2026} introduces additional pose changes in some compositional edits, and RG-DNO~\cite{rgdno2026} shows larger deviations from the source, consistent with the measured  NAD and NJD. Our AGLR modifies the selected limb and torso attributes while largely preserving motion in unedited body parts.

\begin{table}[t]
\centering
\captionsetup{font=footnotesize,skip=3pt}
\renewcommand{\arraystretch}{1.05}
\setlength{\tabcolsep}{1.5pt}
\fontsize{7}{8}\selectfont
\resizebox{\columnwidth}{!}{%
\begin{tabular}{@{}lrrrrrrrr@{}}
\toprule
& \multicolumn{2}{c}{Generation}
& \multicolumn{5}{c}{Editing}
& \multicolumn{1}{c}{No-edit} \\
\cmidrule(lr){2-3}
\cmidrule(lr){4-8}
\cmidrule(l){9-9}
Variant
& TAE$\downarrow$
& ASR$\uparrow$
& TAE$\downarrow$
& NAD$\downarrow$
& NJD$\downarrow$
& ASR$_c^a\uparrow$
& ASR$_c^j\uparrow$
& ID$\downarrow$ \\
\midrule
AGLR (Full)
& \underline{0.0740} & \underline{80.34}
& \underline{0.1431} & \textbf{0.0584}
& \textbf{0.681} & \textbf{76.81}
& \textbf{77.28} & \textbf{0.578} \\
\midrule
w/o base latent
& 0.7647 & 21.81 & 0.5122 & 0.1434
& 2.270 & 24.05 & 19.00 & 1.848 \\

w/o GT superv.
& 0.2716 & 50.40 & 0.2516
& 0.0820 & 0.911
& 55.96 & 56.03 & 0.669 \\

w/o ID superv.
& 0.0762 & 79.76 & 0.1494
& \underline{0.0595} & \underline{0.687}
& \underline{76.26} & \underline{76.78} & 3.236 \\

w/o Edit superv.
& 0.6895 & 30.93 & 0.9832
& 0.0676 & 0.723
& 10.99 & 11.01 & 0.644 \\

w/o attr. preserv.
& 0.0842 & 78.42 & 0.1457 & 0.1587
& 0.876 & 69.37 & 74.71 & 0.663 \\
w/o coord. preserv.
& 0.0799 & 76.83 & 0.1530 & 0.0773
& 3.601 & 72.59 & 34.64
& \underline{0.634} \\
w/o both preserv.
& \textbf{0.0588} & \textbf{85.55}
& \textbf{0.1342} & 0.6637
& 17.425 & 7.79 & 6.01 & 0.635 \\
Full latent pred.
& 0.1393 & 70.67 & 0.1848 & 0.0982
& 1.209 & 72.00 & 72.47 & 1.198 \\
\bottomrule
\end{tabular}%
}
\caption{\textbf{Ablation studies on controllable generation and source-conditioned editing.}
ASR$_c^a$ and ASR$_c^j$ denote success rates under separate
NAD $\leq 0.30$ and NJD $\leq 3^\circ$ constraints, respectively. ID measures no-edit joint RMSE. Joint errors are reported in degrees.}
\label{tab:ablation}
\end{table}

\subsection{Ablation Studies}
We evaluate each ablation on attribute-conditioned generation and source-conditioned editing, using 3,766 requests per task. To evaluate no-edit consistency, we pass 256 ground-truth motions to each variant without requesting any edits. We report the joint-space RMSE between input and output motions as the ID metric.

\noindent\textbf{Base conditioning and residual prediction.}
In Table~\ref{tab:ablation}, 
removing base-latent conditioning reduces generation ASR
from 80.34\% to 21.81\% and editing ASR$_c^a$ from
76.81\% to 24.05\%. Direct prediction of the refined latent also lowers performance on both tasks. These results support conditioning the refinement on the base latent and predicting an additive residual.

\noindent\textbf{Training supervision.}
Removing GT supervision reduces generation ASR from 80.34\% to 50.40\% and editing \(\mathrm{ASR}_c^a\) from 76.81\% to 55.96\%. Removing Edit supervision causes larger drops, to 30.93\% and 10.99\%, respectively. Removing ID supervision increases no-edit joint RMSE from \(0.578^\circ\) to \(3.236^\circ\), while generation ASR and editing \(\mathrm{ASR}_c^a\) each decrease by less than one percentage point. These results indicate that GT and Edit supervision contribute to target adherence, while ID supervision primarily helps preserve the input motion when no edit is requested.

\noindent\textbf{Complementary preservation losses.}
Removing the attribute preservation loss increases NAD from 0.0584 to 0.1587, while removing the coordinate preservation loss increases NJD from \(0.681^\circ\) to \(3.601^\circ\). Removing both losses raises generation ASR from 80.34\% to 85.55\%, and sharply reduces editing success under both preservation constraints. Editing \(\mathrm{ASR}_c^a\) falls from 76.81\% to 7.79\%, and editing \(\mathrm{ASR}_c^j\) falls from 77.28\% to 6.01\%. These results show that both losses help the model meet editing targets while preserving unedited attributes and joint motion.

\begin{table}[t]
\centering
\captionsetup{font=footnotesize,skip=3pt}
\renewcommand{\arraystretch}{0.95}
\setlength{\tabcolsep}{1.5pt}
\fontsize{7}{8}\selectfont
\resizebox{\columnwidth}{!}{%
\begin{tabular}{@{}llrrrrrrrr@{}}
\toprule
Method & Ref.
& TAE$\downarrow$
& ASR$\uparrow$
& FID$\downarrow$
& R@1$\uparrow$
& R@2$\uparrow$
& R@3$\uparrow$
& MD$\downarrow$
& Div. \\
\midrule
OmniControl~\cite{omnicontrol2024} & ICLR'24
& 0.8052 & 15.97 & 0.3168
& 30.24 & 46.07 & 55.75
& 1.1623 & 1.2868 \\
TLControl~\cite{tlcontrol2024} & ECCV'24
& 0.8648 & 16.08 & 0.6316
& 11.34 & 17.19 & 23.44
& 1.3002 & 1.0768 \\
MotionLCM~\cite{motionlcm2024} & ECCV'24
& 0.4551 & 31.18 & 0.3604
& \underline{42.59} & \underline{57.51}
& \underline{66.48} & \underline{1.0757} & 1.2118 \\
MaskControl~\cite{maskcontrol2025} & ICCV'25
& 0.7688 & 17.50 & \underline{0.2669}
& 39.72 & 54.79 & 62.90
& 1.1023 & 1.2990 \\
SFControl~\cite{sfcontrol2025} & ICCV'25
& 0.2933 & 44.47 & 0.6186
& 24.14 & 30.80 & 36.49
& 1.2285 & 1.0613 \\
ProjFlow~\cite{projflow2026} & CVPR'26
& \underline{0.1003} & \underline{74.50}
& 0.2708 & 41.03 & 55.49 & 64.62
& 1.1007 & 1.2850 \\
\midrule
AGLR (ours) & ---
& \textbf{0.0947} & \textbf{75.61}
& \textbf{0.2312} & \textbf{47.98}
& \textbf{62.55} & \textbf{70.51}
& \textbf{1.0126} & 1.2664 \\
\bottomrule
\end{tabular}%
}
\caption{\textbf{Zero-shot fine-grained controllable generation
on BONES-SEED-G1.} Models trained on HumanML3D-G1 are evaluated on
2,000 motions using text and selected GT attributes,
without fine-tuning.}
\vspace{-10pt}
\label{tab:bones_generation}
\end{table}

\subsection{Zero-Shot Transfer}
\label{sec:zero-shot}
We evaluate attribute-conditioned generation and source-conditioned editing on BONES-SEED-G1 using the HumanML3D-G1-trained weights of all compared methods, without fine-tuning. To assess generation quality, we use a frozen TMR evaluator pretrained on the BONES-SEED-G1 training split. 
As shown in Table~\ref{tab:bones_generation} and Table~\ref{tab:bones_editing},
AGLR achieves the lowest generation TAE, the highest generation ASR, and the highest editing success rates under both attribute and joint preservation constraints. These results show that AGLR transfers fine-grained geometric control and selective motion editing across datasets without fine-tuning.

\begin{table}[h]
\centering
\captionsetup{font=footnotesize,skip=3pt}
\renewcommand{\arraystretch}{0.95}
\setlength{\tabcolsep}{2pt}
\fontsize{7}{8}\selectfont
\begin{tabular*}{\columnwidth}{
@{\extracolsep{\fill}}llrrrrr@{}}
\toprule
Method & Ref.
& TAE$\downarrow$
& NAD$\downarrow$
& NJD$\downarrow$
& ASR$_c^a$$\uparrow$
& ASR$_c^j$$\uparrow$ \\
\midrule
TRACE~\cite{trace2026} & arXiv'26
& 0.2370 & \textbf{0.0560}
& \textbf{0.000} & 61.75 & 62.32 \\
OmniME~\cite{omnime2026} & CVPR'26
& 0.2430 & 0.1255
& 1.664 & \underline{61.85} & \underline{62.65} \\
ScaleMoGen~\cite{scalemogen2026} & ECCV'26
& 0.9673 & 0.4686
& 10.007 & 7.72 & 2.59 \\
DNO~\cite{dno2024} & CVPR'24
& 0.1742 & 0.6588
& 14.205 & 6.78 & 0.00 \\
RG-DNO~\cite{rgdno2026} & CVPR'26
& \textbf{0.1144} & 1.2847
& 30.362 & 0.00 & 0.00 \\
\midrule
AGLR (ours) & ---
& \underline{0.1663} & \underline{0.0726}
& \underline{1.106} & \textbf{74.28} & \textbf{74.35} \\
\bottomrule
\end{tabular*}
\caption{\textbf{Zero-shot source-conditioned motion editing
on BONES-SEED-G1.}
Models trained on HumanML3D-G1 edit the same
2,000 GT source motions without fine-tuning.
ASR$_c^a$ and ASR$_c^j$ use separate NAD $\leq 0.30$
and NJD $\leq 3^\circ$ constraints, respectively.}
\label{tab:bones_editing}
\vspace{-10pt}
\end{table}

\subsection{Tracking and Visualization}
\begin{figure}[t]
    \centering
    \includegraphics[width=\columnwidth]{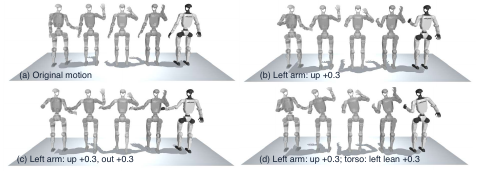}
    \caption{\textbf{Tracking of edited motions.} Robot poses are shown at 0.5, 1.5, 3.0, 4.5, and 5.5\,s, darker poses indicating later frames.}
    \label{fig:tracking}
    \vspace{-10pt}
\end{figure}

We use the edited motions as references for a tracking policy. Fig.~\ref{fig:tracking} compares the original motion with three edited executions at five time points. The robot raises its left arm in Fig. 5(b), raises and extends it outward in Fig. 5(c), and combines the arm raise with a leftward torso lean in Fig. 5(d). Across three edits, the right arm follows a similar pattern to the original, and the lower-body poses remain visually similar at the sampled time points. These examples illustrate that the robot can execute attribute edits while largely preserving the motion of unedited body parts. Additional real-world demonstrations are provided in the supplement.

\section{Conclusion}


We presented AGLR, a unified framework for fine-grained humanoid motion generation and editing that enables compositional, time-varying control over limb extents, torso orientation, and root trajectories. The Attribute Residual Adapter refines a generated or encoded motion latent in a single pass. Attribute- and motion-level supervision jointly guide the model to match target attributes while preserving unselected attributes and untargeted motion coordinates. Experiments show that AGLR compares favorably with adapted baselines in control accuracy and motion preservation, supports efficient editing, and transfers across motion sources without fine-tuning.
Limitations are discussed in the supplement.



{
    \small
    \bibliographystyle{ieeenat_fullname}
    \bibliography{ref}
}

\clearpage
\maketitlesupplementary
\appendix

\renewcommand{\topfraction}{0.95}
\renewcommand{\dbltopfraction}{0.95}
\renewcommand{\textfraction}{0.05}
\renewcommand{\floatpagefraction}{0.8}
\renewcommand{\dblfloatpagefraction}{0.8}

\setcounter{section}{0}
\setcounter{figure}{0}
\setcounter{table}{0}
\setcounter{equation}{0}
\renewcommand{\thesection}{\Alph{section}}
\renewcommand{\theHsection}{A\arabic{section}}
\renewcommand{\thefigure}{S\arabic{figure}}
\renewcommand{\theHfigure}{S\arabic{figure}}
\renewcommand{\thetable}{S\arabic{table}}
\renewcommand{\theHtable}{S\arabic{table}}
\renewcommand{\theequation}{S\arabic{equation}}
\renewcommand{\theHequation}{S\arabic{equation}}

\emergencystretch=0.5em
\parfillskip=0pt plus 0.5\linewidth
\clubpenalty=1000
\widowpenalty=1000
\displaywidowpenalty=1000

This supplementary material provides additional results and implementation details for AGLR. Section \ref{sec:supp-real-world} presents real-world motion editing and tracking demonstrations on the Unitree G1 humanoid robot. Section \ref{sec:WGAR} provides implementation details and a local controllability analysis of WGAR. Sections \ref{sec:C}–\ref{sec:E} describe the architectures and training procedures of the motion VAE, motion generator, and ARA, respectively. Section \ref{sec:supp-asr-analysis} analyzes sensitivity to preservation thresholds, and Section \ref{sec:F} details the editing latency measurement protocol. Section \ref{sec:supp-control-benchmark} describes baseline adaptations and training settings. Finally, Section \ref{sec:supp-limitations} discusses limitations.
Additional videos of robot execution and an interactive editing demonstration are available on our project page.

\section{Real-World Motion Editing and Tracking}
\label{sec:supp-real-world}
We demonstrate attribute-guided motion editing on a physical Unitree G1 humanoid robot. Figures~\ref{fig:single_attribute_arm_up}--\ref{fig:root_orientation} 
show directional limb adjustments, simultaneous control of multiple body parts, editing with time-varying targets, and modifications to root trajectories and heading. Each figure compares execution of the base reference motion with executions of its edited variants.

\noindent
\textbf{From Geometric Requests to Robot Execution.}
Given a base motion $\mathbf X_b$, we encode it into a latent representation $\mathbf z_b=E(\mathbf X_b)$ and provide ARA with the selected attribute targets over time. For an offset-based request, we define the target as $A_{t,k}=A_{b,t,k}+\delta_k(t)$, where  $\mathbf A_b=\operatorname{WGAR}(\mathbf X_b)$ and \(\delta_k(t)\) denotes the requested offset for attribute \(k\) at time \(t\). Each request jointly specifies all active targets. Repeated requests for the same base motion reuse the cached latent representation $\mathbf z_b$. The resulting edited reference is executed on the physical robot using the PPO-based tracking policy from MimicLite~\cite{mimiclite2026}, conditioned on the robot's current state.

\noindent
\textbf{Results.}
Figures~\ref{fig:single_attribute_arm_up}--\ref{fig:root_orientation} illustrate 
editing results using three base motions: Base 0 for limb, torso, and time-varying attribute editing; Base 1 for root trajectory editing; and Base 2 for turning while marching. Limb offsets are dimensionless, while torso offsets are expressed in radians. Arm and leg directions are defined in the torso and pelvis frames, respectively, with positive Out pointing away from the body on both sides. Root yaw editing activates all three root-rotation channels, and the \(180^\circ\) and \(360^\circ\) turning requests specify full heading trajectories over time. These qualitative results demonstrate that the edited references can be executed on the physical robot to produce the requested geometric changes while preserving the overall structure of the base motions.

\section{Implementation Details of WGAR}
\label{sec:WGAR}
\subsection{Geometric Attribute Details}
We provide additional implementation details for WGAR, introduced in Section 3.1 of the main paper. Table~\ref{tab:attribute_channels} summarizes the channel ordering, link selection, neutral alignment, and limb-length normalization. Limb attributes are dimensionless, positional attributes are expressed in metres, and angular attributes are expressed in radians. As defined in the main paper, limb attributes are computed in the torso or pelvis frame, torso attributes in the pelvis frame, and root displacement and orientation in the initial-heading frame. 

\begin{table}[t]
\centering
\caption{Channel ordering of the Whole-body Geometric Attributes Representation (WGAR). Channel indices start from zero.}
\label{tab:attribute_channels}
\small
\begin{tabular}{@{}lp{0.75\linewidth}@{}}
\toprule
Channels & Attributes \\
\midrule
0--2   & Left arm: Front, Out, Up \\
3--5   & Left leg: Front, Out, Up \\
6--8   & Right arm: Front, Out, Up \\
9--11  & Right leg: Front, Out, Up \\
12--14 & Torso: Forward Tilt, Rightward Tilt, Heading \\
15--16 & Root horizontal displacement in the initial-heading frame \\
17     & Absolute root height \\
18--20 & Root roll, pitch, and yaw in the initial-heading frame \\
\bottomrule
\end{tabular}
\end{table}

\paragraph{Limb Link Selection} Table \ref{tab:limb_links} lists the anchor, middle, and endpoint links for each limb \(k\), indexed by \(a_k\), \(m_k\), and \(e_k\), respectively. For each arm, these correspond to the shoulder, elbow, and wrist links; for each leg, they correspond to the hip, knee, and ankle links. Each link name uses the corresponding \texttt{left\_} or \texttt{right\_} prefix. For limb \(k\), their positions, \(\mathbf p_{t,a_k}\), \(\mathbf p_{t,m_k}\), and \(\mathbf p_{t,e_k}\), are obtained through forward kinematics and used to compute the limb attributes.
\begin{table}[t]
\centering
\caption{Links used to compute limb attributes. Each link name is prefixed with \texttt{left\_} or \texttt{right\_}.}
\label{tab:limb_links}
\scriptsize
\setlength{\tabcolsep}{3pt}
\begin{tabular}{@{}llll@{}}
\toprule
Limb & Anchor & Middle joint & Endpoint \\
\midrule
Arm & \texttt{shoulder\_pitch\_link}
    & \texttt{elbow\_link}
    & \texttt{wrist\_yaw\_link} \\
Leg & \texttt{hip\_pitch\_link}
    & \texttt{knee\_link}
    & \texttt{ankle\_roll\_link} \\
\bottomrule
\end{tabular}
\end{table}

\paragraph{Neutral Alignment and Limb-Length Normalization}
Let \(\mathbf q_t\) denote the joint configuration at frame \(t\). For each limb \(k\), let \(b_k\) denote its reference body link: \texttt{torso\_link} for arms and \texttt{pelvis} for legs. We denote the global orientation of this link at frame \(t\) by \(\mathbf R_{t,b_k}\), and its orientation in the rest pose by \(\mathbf R^{0}_{b_k}\). The neutral-aligned orientation used in Eq. (1) of the main paper is then defined as:
\begin{equation}
\mathbf Q_{t,k}
=
\mathbf R_{t,b_k}
\left(\mathbf R^{0}_{b_k}\right)^\top.
\end{equation}
Here, \(\mathbf R^{0}_{b_k}\) is computed by forward kinematics (FK) with zero joint angles and an identity root orientation. This alignment removes the link's rest-pose rotational offset, yielding an identity orientation in the rest pose while preserving subsequent global rotations.

For limb-length normalization, let \(\mathbf p^{0}_{a_k}\), \(\mathbf p^{0}_{m_k}\), and \(\mathbf p^{0}_{e_k}\) denote the rest-pose positions of the anchor, middle, and endpoint links, respectively. We define the normalization factor as the sum of the two segment lengths:
\begin{equation}
\ell_k
=
\left\|\mathbf p^{0}_{m_k}-\mathbf p^{0}_{a_k}\right\|_2
+
\left\|\mathbf p^{0}_{e_k}-\mathbf p^{0}_{m_k}\right\|_2.
\label{eqS2}
\end{equation}
The reference rotations \(\mathbf R^{0}_{b_k}\) and limb lengths \(\ell_k\) are computed once from the same rest pose and held fixed across all motions.

\paragraph{Orientation Conventions}
The operator \(\operatorname{rpy}(\cdot)\) in Eq. (3) of the main paper returns roll, pitch, and yaw angles under the ZYX Euler-angle convention, such that
$\mathbf R=
\mathbf R_z(\mathrm{yaw})
\mathbf R_y(\mathrm{pitch})
\mathbf R_x(\mathrm{roll})$.
Angular trajectories are temporally unwrapped to maintain continuity across angle wrap boundaries, allowing heading targets to represent turns beyond a single revolution.

\subsection{Editable Motion Variables and Preservation}
Table \ref{tab:editable_variables} lists the motion variables in raw-space representation (RPR39) that can be modified for each selected attribute. The same selection applies to all frames in the sequence. When multiple attributes are selected, a motion variable is editable if it belongs to the editable set of any selected attribute. Selecting root orientation also allows root displacement to change, since displacement increments are expressed in a heading-local frame and therefore depend on the root heading.

We encourage preservation at both the motion-coordinate and attribute levels. The coordinate-preservation objective encourages all motion variables outside the editable set to remain close to the base motion, with differences measured in standardized RPR39 space. The attribute-preservation objective encourages unselected attributes to retain their base values, including those that depend on editable motion variables. For example, selecting an arm's Up attribute allows the corresponding arm joints to change while encouraging its Front and Out attributes to remain unchanged. 
Attribute preservation compares the edited and base motions in WGAR space, with each attribute computed in its corresponding reference frame as defined above.

\begin{table}[t]
\centering
\caption{Mapping from selected attributes to editable motion-space variables. Left/right limb selections affect only the corresponding limb.}
\label{tab:editable_variables}
\small
\setlength{\tabcolsep}{4pt}
\renewcommand{\arraystretch}{1.15}
\begin{tabular}{@{}p{0.32\linewidth}p{0.62\linewidth}@{}}
\toprule
Selected attributes & Editable motion-space variables \\
\midrule
Any arm attribute
& Shoulder, elbow, and wrist joint angles of the corresponding arm \\
Any leg attribute
& Hip, knee, and ankle joint angles of the corresponding leg \\
Any torso attribute
& Waist joint angles \\
Root horizontal displacement
& Two heading-local horizontal displacement increments \\
Root orientation
& Two heading-local horizontal displacement increments, yaw increment, and 6D heading-local rotation representation \\
Root height
& Root height \\
\bottomrule
\end{tabular}
\end{table}

\subsection{Local Controllability of Limb Attributes}
\label{sec:attribute_coupling}
Limb attributes depend on the same joint angles and are therefore coupled: raising an arm may also change its forward and outward extent. However, adjusting several joints together can sometimes compensate for these accompanying changes. We examine when it is locally possible to increase Up while keeping Front and Out unchanged.

The limb attribute vector in Eq. (1) of the main paper contains three directional components:
\begin{equation}
\mathbf a_{t,k}
=
[\mathrm{Front}_{t,k},\mathrm{Out}_{t,k},\mathrm{Up}_{t,k}]^\top.
\end{equation}
For clarity, we omit the frame index \(t\) below. With the anchor position and reference body frame held fixed, we write \(\mathbf a_k=f_k(\mathbf q_k)\), where \(\mathbf q_k\in\mathbb R^{n_k}\) contains the editable joint angles. A first-order Taylor expansion around the current configuration gives:
\begin{equation}
\begin{aligned}
\Delta\mathbf a_k
&=f_k(\mathbf q_k+\Delta\mathbf q_k)-f_k(\mathbf q_k)\\
&=\mathbf J_k\Delta\mathbf q_k
+O(\|\Delta\mathbf q_k\|_2^2),
\end{aligned}
\end{equation}
where \(\mathbf J_k=\partial f_k/\partial\mathbf q_k\in\mathbb R^{3\times n_k}\) is the attribute Jacobian. Its three rows describe how joint adjustments affect Front, Out, and Up, respectively. For sufficiently small joint changes, the higher-order term can be neglected, giving \(\Delta\mathbf a_k\approx\mathbf J_k\Delta\mathbf q_k\).
To increase Up by \(\delta\) while preserving Front and Out, we seek a joint adjustment satisfying:
\begin{equation}
\mathbf J_k\Delta\mathbf q_k
=
\Delta\mathbf a_k^{\mathrm{target}}
=
\begin{bmatrix}0\\0\\\delta\end{bmatrix}.
\end{equation}
If \(\mathbf J_k\) has full row rank, all three attribute directions are independently controllable in this local linear approximation. One solution is:
\begin{equation}
\Delta\mathbf q_k
=
\mathbf J_k^\dagger\Delta\mathbf a_k^{\mathrm{target}},
\end{equation}
where \(\dagger\) denotes the Moore–Penrose pseudoinverse. Under this rank condition, \(\mathbf J_k\mathbf J_k^\dagger=\mathbf I_3\), so the update produces exactly the target change in the linearized model. In other words, coordinated joint adjustments can cancel changes in Front and Out while increasing Up.


For the nonlinear attribute mapping, a single such update may leave small higher-order deviations. Nevertheless, assuming the mapping is smooth, sufficiently small target attribute changes are locally achievable when the current configuration lies strictly within the joint limits and the Jacobian has full row rank.


\section{Motion VAE Architecture and Training}
\label{sec:C}
\subsection{VAE Architecture}
The motion VAE uses a temporal encoder–decoder architecture with approximately 15.5M parameters. The encoder and decoder contain three and two Transformer blocks, respectively, operating at the compressed temporal resolution. Each block uses pre-layer normalization, eight-head self-attention, and a residual feed-forward network with hidden dimensions $384\!\rightarrow\!1536\!\rightarrow\!384$ and GELU activation~\cite{hendrycks2016gelu}. Separate residual temporal-convolution blocks accompany the downsampling and upsampling stages. The encoder downsamples the motion sequence in two stages, reducing its length by a factor of four, and maps the resulting features to 64-dimensional latent tokens. Thus, a 500-frame motion sequence is encoded as 125 latent tokens. The decoder reconstructs the motion from these tokens through two temporal upsampling stages, restoring the original sequence length and motion feature dimension.

\subsection{Posterior and Deterministic Encoding}
Let \(\mathbf X\in\mathbb R^{T\times d_x}\) denote a motion sequence with \(T\) valid frames and feature dimension \(d_x=39\). Before encoding, we standardize each feature using fixed training-set statistics: \(\mathbf x=\mathcal N_x(\mathbf X)\). The encoder maps \(\mathbf x\) to the mean \(\boldsymbol\mu_\phi(\mathbf x)\) and log variance \(\log\boldsymbol\sigma_\phi^2(\mathbf x)\) of a diagonal Gaussian posterior. Both outputs have shape \(L\times d_z\), where \(L=\lceil T/4\rceil\) is the latent sequence length and \(d_z=64\) is the dimension of each latent token. 
During VAE training, we sample a latent representation using the reparameterization trick and decode it to reconstruct the standardized motion:
\begin{equation}
\begin{aligned}
q_\phi(\mathbf h\mid\mathbf x)
&=
\mathcal N\!\left(
\boldsymbol\mu_\phi(\mathbf x),
\operatorname{diag}\!\big(\boldsymbol\sigma_\phi^2(\mathbf x)\big)
\right),\\
\mathbf h
&=
\boldsymbol\mu_\phi(\mathbf x)
+\boldsymbol\sigma_\phi(\mathbf x)\odot\boldsymbol\eta,
\qquad
\eta_{i,j}\overset{\mathrm{i.i.d.}}{\sim}\mathcal N(0,1),\\
\hat{\mathbf x}
&=D_{\mathrm{raw}}(\mathbf h),
\end{aligned}
\end{equation}
Here, \(\mathbf h\in\mathbb R^{L\times d_z}\), \(\odot\) denotes element-wise multiplication, and \(D_{\mathrm{raw}}\) outputs motion features in standardized space. The Gaussian posterior is defined over the flattened latent entries, with a separate variance for each entry. 

For flow training and ARA, we use the posterior mean instead of sampling, so that each motion sequence has a deterministic encoding. We further standardize these latent representations using fixed training-set statistics, denoted by \(\mathcal N_z\). The encoder mapping used in the main paper is therefore
\begin{equation}
E(\mathbf X)
=
\mathcal N_z\!\left(
\boldsymbol\mu_\phi\!\left(\mathcal N_x(\mathbf X)\right)
\right),
\end{equation}
To decode a standardized latent representation \(\mathbf z\), we first undo latent standardization, apply the VAE decoder, and then undo motion standardization:
\begin{equation}
D(\mathbf z)
=
\mathcal N_x^{-1}\!\left(
D_{\mathrm{raw}}\!\left(\mathcal N_z^{-1}(\mathbf z)\right)
\right).
\end{equation}
Thus, \(E\) maps motion sequences to standardized latent representations, while \(D\) reconstructs motion variables in their original units.

\subsection{Reconstruction and Regularization Objectives}
We train the VAE to reconstruct both standardized RPR39 features and Cartesian body positions computed through FK. We also supervise temporal differences to preserve motion dynamics and regularize the latent posterior toward a standard Gaussian. The overall objective is:
\begin{equation}
\begin{aligned}
\mathcal L_{\mathrm{VAE}}
={}&\mathcal L_{\mathrm{rec}}
+\lambda_{\mathrm{vel}}\mathcal L_{\mathrm{vel}}
+\lambda_{\mathrm{acc}}\mathcal L_{\mathrm{acc}}\\
&+\lambda_{\mathrm{FK}}\mathcal L_{\mathrm{FK}}
+\beta\mathcal L_{\mathrm{KL}},
\end{aligned}
\end{equation}
where \(\mathcal L_{\mathrm{rec}}\) measures motion-feature reconstruction error, \(\mathcal L_{\mathrm{vel}}\) and \(\mathcal L_{\mathrm{acc}}\) measure errors in first- and second-order temporal differences, \(\mathcal L_{\mathrm{FK}}\) measures body-position reconstruction error, and \(\mathcal L_{\mathrm{KL}}\) regularizes the posterior. We set \(\lambda_{\mathrm{vel}}=0.5\), \(\lambda_{\mathrm{acc}}=0.25\), and \(\lambda_{\mathrm{FK}}=1\). The KL weight \(\beta\) is gradually increased to \(10^{-4}\) over the first 20 epochs.

\paragraph{Motion Feature Reconstruction} 
Let \(\mathbf x\) and \(\hat{\mathbf x}\) denote the standardized target and reconstructed motion sequences, respectively. We supervise the motion features and their first- and second-order temporal differences using:
\begin{equation}
\begin{aligned}
(\Delta^0\mathbf x)_t
&=\mathbf x_t,\\
(\Delta^1\mathbf x)_t
&=\mathbf x_{t+1}-\mathbf x_t,\\
(\Delta^2\mathbf x)_t
&=\mathbf x_{t+2}-2\mathbf x_{t+1}+\mathbf x_t.
\end{aligned}
\end{equation}
These three quantities describe the feature values, their frame-to-frame changes, and how those changes vary over time. For each order \(o\in\{0,1,2\}\), we define:
\begin{equation}
\mathcal L_o
=
\frac{1}{|\mathcal V_o|}
\sum_{(t,j)\in\mathcal V_o}
\rho\!\left(
(\Delta^o\hat{\mathbf x})_{t,j}
-
(\Delta^o\mathbf x)_{t,j}
\right),
\end{equation}
where \(j\) indexes motion features, \(\mathcal V_o\) contains the valid frame-feature entries for order \(o\), and $\rho$ denotes the smooth L1 penalty. The reconstruction and temporal losses are
\begin{equation}
\mathcal L_{\mathrm{rec}}=\mathcal L_0,\qquad
\mathcal L_{\mathrm{vel}}=\mathcal L_1,\qquad
\mathcal L_{\mathrm{acc}}=\mathcal L_2.
\end{equation}

\paragraph{FK Position Reconstruction} 
To supervise body geometry, we first convert the reconstructed features back to their original units: \(\hat{\mathbf X}=\mathcal N_x^{-1}(\hat{\mathbf x})\). We then apply FK to the target and reconstructed motions to obtain body positions:
\begin{equation}
\mathbf p_{t,b}
=
\operatorname{FK}(\mathbf X)_{t,b},
\qquad
\hat{\mathbf p}_{t,b}
=
\operatorname{FK}(\hat{\mathbf X})_{t,b},
\end{equation}
where \(\mathbf p_{t,b}\) and \(\hat{\mathbf p}_{t,b}\) denote the target and reconstructed positions of body \(b\) at frame \(t\), measured in metres. The position reconstruction loss is:
\begin{equation}
\mathcal L_{\mathrm{FK}}
=
\frac{1}{3T|\mathcal B|}
\sum_{t=1}^{T}
\sum_{b\in\mathcal B}
\sum_{d=1}^{3}
\rho\!\left(\hat p_{t,b,d}-p_{t,b,d}\right),
\end{equation}
where \(T\) is the number of valid frames, \(d\) indexes the three Cartesian coordinates, and \(\mathcal B\) includes the free-root body and the bodies associated with articulated joints in the G1 kinematic model.

\paragraph{Posterior Regularization}
We use the KL divergence to encourage the diagonal Gaussian posterior to remain close to a standard Gaussian: 
\begin{equation}
\mathcal L_{\mathrm{KL}}
=
-\frac{1}{2|\mathcal V_z|}
\sum_{(i,j)\in\mathcal V_z}
\left(
1+\log\sigma_{i,j}^{2}
-\mu_{i,j}^{2}
-\sigma_{i,j}^{2}
\right),
\end{equation}
where \(\mu_{i,j}\) and \(\sigma_{i,j}\) are the posterior mean and standard deviation for latent token \(i\) and feature \(j\), and \(\mathcal V_z\) contains valid latent-token/feature entries. 

\subsection{Training Details}

We train the VAE on the HumanML3D-G1 training split for 800 epochs, using motion sequences of up to 500 frames. The global batch size is 2,048. We use the AdamW optimizer with a target learning rate of \(10^{-4}\), betas \((0.9,0.99)\), and zero weight decay. The learning rate increases to \(10^{-4}\) over the first five epochs and then follows a cosine decay schedule to \(10^{-6}\). Gradients are clipped at a norm of 1. 

\section{Generator Architecture and Training}

\subsection{Teacher Motion Generator}
\label{sec:supp-flow-architecture}
The motion generator is a text-conditioned latent flow model that operates in the latent space of the frozen motion VAE. Given a noisy motion latent \(\mathbf z_\tau\), flow time \(\tau\), text condition \(c\), and motion duration \(T\), the model predicts a latent velocity with the same shape as the input, \(L\times d_z\). During generation, this velocity field progressively transforms Gaussian noise into a motion latent, which is then decoded using the frozen VAE decoder.

\paragraph{Generator Architecture} 
The generator uses a Transformer backbone with 12 blocks, a hidden dimension of 768, and 12 attention heads. An input MLP projects the noisy latent tokens into the hidden feature space, while learned positional embeddings encode their temporal positions. Each Transformer block contains bidirectional self-attention, cross-attention to the conditioning tokens, and a feed-forward network with an expansion ratio of four and GELU activation. Flow time modulates the hidden features through adaptive layer normalization. After the Transformer blocks, a time-conditioned output head maps the hidden features to latent velocities.

\paragraph{Conditioning Mechanisms} 
Text features are extracted using a frozen CLIP ViT-B/32~\cite{radford2021clip} encoder and projected to the Transformer hidden dimension through a trainable two-layer MLP. Motion duration is encoded through a sinusoidal embedding followed by an MLP to produce a single duration token. We concatenate the text tokens and duration token along the token dimension to form the cross-attention context:
\begin{equation}
\mathbf C_{\mathrm{flow}}
=
\operatorname{Concat}
\left(\mathbf C_{\mathrm{text}},\mathbf c_{\mathrm{len}}\right),
\end{equation}
Flow time is embedded separately and used for adaptive layer normalization. Padding masks exclude invalid motion and text tokens from attention.

\paragraph{Flow Training Objective}
With the VAE frozen, we encode each training motion as a standardized clean latent \(\mathbf z_{\mathrm{GT}}=E(\mathbf X_{\mathrm{GT}})\). We construct a linear interpolation between the clean latent and Gaussian noise:
\begin{equation}
\mathbf z_\tau
=
(1-\tau)\mathbf z_{\mathrm{GT}}
+\tau\boldsymbol\epsilon,
\qquad
\boldsymbol\epsilon\sim\mathcal N(\mathbf 0,\mathbf I).
\end{equation}
Here, \(\tau=0\) corresponds to the clean motion latent and \(\tau=1\) to noise. We sample \(\tau\sim\operatorname{Beta}(1.5,1)\), assigning greater probability to the noisier portion of the path. The flow model is trained to predict the velocity along this path using
\[
\mathcal L_{\mathrm{FM}}
=
\mathbb E\!\left[
\frac{1}{|\mathcal V|}
\sum_{(\ell,h)\in\mathcal V}
\left(
\left[v_\theta(\mathbf z_\tau,\tau,c,T)\right]_{\ell,h}
-
\left[\boldsymbol\epsilon-\mathbf z_{\mathrm{GT}}\right]_{\ell,h}
\right)^2
\right].
\]
where \(\mathcal V\) contains the valid latent entries, with \(\ell\) indexing tokens and \(h\) indexing features. The expectation is over training motion–text pairs, temporal crops, Gaussian noise, and flow time. 

\paragraph{Teacher sampling}
At inference, we start from Gaussian noise \(\mathbf z_1\sim\mathcal N(\mathbf 0,\mathbf I)\) and integrate the learned velocity field backward from \(\tau=1\) to \(\tau=0\) using 50 Euler steps of equal size. At each step, classifier-free guidance combines the text-conditioned and empty-text velocity predictions:
\begin{equation}
\begin{aligned}
\mathbf v_c
&=v_\theta(\mathbf z_{\tau_k},\tau_k,c,T),\\
\mathbf v_\varnothing
&=v_\theta(\mathbf z_{\tau_k},\tau_k,\varnothing,T),\\
\mathbf v_{\mathrm{cfg}}
&=\mathbf v_\varnothing+s_{\mathrm{cfg}}(\mathbf v_c-\mathbf v_\varnothing),\\
\mathbf z_{\tau_{k+1}}
&=\mathbf z_{\tau_k}
+(\tau_{k+1}-\tau_k)\mathbf v_{\mathrm{cfg}},
\end{aligned}
\end{equation}
where \(\tau_k=1-k/50\) for \(k=0,\ldots,50\), and the guidance scale is set to \(s_{\mathrm{cfg}}=3\). Both predictions retain the same duration condition \(T\); only the text condition differs. Each Euler step requires two velocity predictions, giving 100 velocity-network evaluations per generated sequence. Finally, the frozen VAE decoder maps \(\mathbf z_0\) back to motion through \(D\).

\subsection{Few-Step Distillation}\label{sec:supp-distillation}
We adopt DMD2~\cite{yin2024dmd2} to distill the flow teacher into a four-step motion generator using distribution-matching and adversarial objectives. Both the student and the auxiliary fake-distribution network use the teacher's architecture and are initialized from its exponential moving average (EMA) weights. The teacher, CLIP encoder, and text projector remain frozen throughout distillation.

\paragraph{Distillation Setup}
Our flow model predicts latent velocities, whereas the distribution-matching objective uses clean-latent estimates. Under the linear interpolation path, we convert a velocity prediction \(\mathbf v\) at flow time \(\tau\) into a clean-latent estimate as $\tilde{\mathbf z}=\mathbf z_\tau-\tau\mathbf v$.
We use the estimates from the teacher and auxiliary network to construct the distribution-matching objective. For adversarial training, we attach a discriminator head to the auxiliary network. Its final-block features are averaged over valid tokens, then passed through LayerNorm and a \(768\!\rightarrow\!256\!\rightarrow\!1\) MLP with SiLU activation to produce a scalar discriminator output.

\paragraph{Four-step sampler}
Starting from Gaussian noise \(\mathbf z^{(0)}\sim\mathcal N(\mathbf 0,\mathbf I)\), 
the student generates a motion latent in four stages using the time schedule  \((\tau_0,\tau_1,\tau_2,\tau_3)=(1,0.75,0.5,0.25)\). At each stage \(k\), the student predicts a velocity and converts it into a clean-latent estimate:
\begin{equation}
\mathbf y^{(k)}=
\mathbf z^{(k)}
-\tau_k v_S(\mathbf z^{(k)},\tau_k,c,T),
\ k=0,\ldots,3,
\end{equation}
where \(v_S\) denotes the student velocity network. For the first three stages, we add fresh Gaussian noise to this estimate to construct the input at the next time level:
\begin{equation}
\mathbf z^{(k+1)}
=
(1-\tau_{k+1})\mathbf y^{(k)}
+\tau_{k+1}\boldsymbol\epsilon_k,
\qquad
\boldsymbol\epsilon_k\sim\mathcal N(\mathbf 0,\mathbf I),
\end{equation}
with noise sampled independently at each stage. The fourth-stage estimate, \(\mathbf y^{(3)}\), is the final motion latent and is not re-noised.

Teacher guidance is incorporated during distillation, so student sampling requires no additional empty-text predictions. Generating each sequence therefore requires only four student-network evaluations. For attribute-controlled generation, ARA edits the resulting latent before VAE decoding. For text-only generation, the latent is decoded directly.

\subsection{Training Details}

\paragraph{Teacher Training}
We train the latent flow teacher for 80,000 iterations while keeping the VAE frozen. We use AdamW with betas \((0.9,0.99)\) and zero weight decay, and clip the gradient norm at 1. The learning rate increases to \(10^{-4}\) over the first 1,000 updates and remains constant thereafter. The global batch size is 128 across four GPUs, with 32 samples per GPU. For evaluation, we use exponential moving average (EMA) weights with a decay of 0.9999.

\paragraph{Distillation Training}
We train the student for 10,000 updates, performing five auxiliary-network updates for each student update. The flow times used to evaluate the distribution-matching objective are sampled uniformly from \([0.02,0.98]\). We set the weight of the student's adversarial loss to 0.03 and that of the auxiliary network's discriminator loss to 1.
Both the student and auxiliary network use AdamW with betas \((0.9,0.99)\) and zero weight decay. Their learning rates increase to \(10^{-5}\) over the first 100 updates, and their gradient norms are clipped at 1. The global batch size is 128 across four GPUs, with 32 samples per GPU. We maintain student EMA weights with a decay of 0.999.
For each student update, we uniformly select one of the four sampling stages for backpropagation. The preceding stages generate its input without gradient tracking, so gradients propagate only through the selected stage.

\section{ARA Implementation and Training}
\label{sec:E}
\subsection{ARA Architecture and Optimization}
\paragraph{ARA Architecture}
ARA predicts a latent residual from the base motion latent and the requested attribute controls. It first encodes these inputs into motion tokens and control tokens, respectively. A Transformer then updates the motion tokens through cross-attention to the control tokens. The updated motion tokens are mapped to a residual, which is added to the base latent before decoding with the frozen VAE decoder.

The attribute encoder takes a 44-dimensional vector at each frame, consisting of 21 masked standardized target values, 21 selection-mask entries, a normalized frame index \(t/(T-1)\), and a duration feature \((T-1)/500\). The selection mask is shared across frames. Three temporal convolutional layers with channel dimensions $44\!\rightarrow\!256\!\rightarrow\!256\!\rightarrow\!768$ and GELU activations~\cite{hendrycks2016gelu} produce token embeddings. Two stride-2 layers reduce the temporal resolution by a factor of four to match the motion latent sequence. Invalid temporal positions are masked at each resolution.

The base latent is projected to 768-dimensional motion tokens, to which learned positional embeddings and a duration embedding are added. These motion tokens pass through eight Transformer blocks, each containing 12-head self-attention, cross-attention, and a two-layer feed-forward network. Self-attention captures temporal relationships among motion tokens, while cross-attention uses motion tokens as queries and control tokens as keys and values to incorporate the requested edits.

Finally, layer normalization and an output projection map the updated motion tokens to a residual with the same shape as the base latent. The output projection is initialized to zero, so ARA initially leaves the base latent unchanged.

\paragraph{Attribute Error Definitions}
We define the channel-wise error function \(e\) used in Eq. (5) of the main paper. Channel indices are zero-based and follow the ordering in Table \ref{tab:attribute_channels}. At each valid frame, let \(\hat{\mathbf a},\mathbf a\in\mathbb R^{21}\) denote the attributes of the edited motion and their reference values, respectively. For selected channels, the reference is the requested target; for unselected channels, it is the corresponding base-motion attribute.

To account for differences in attribute scale, we normalize errors using $\sigma_j=\max(s_{A,j},0.05)$,
where \(s_{A,j}\) is the training-set standard deviation of channel \(j\).

For limb attributes, root horizontal displacement, and root height, we use normalized squared errors:
\begin{equation}
e_j(\hat{\mathbf a},\mathbf a)
=
\frac{(\hat a_j-a_j)^2}{\sigma_j^2},
\qquad
j\in\{0,\ldots,11,15,16,17\}.
\end{equation}
For torso attributes, we account for angular periodicity using the shortest signed angular difference:
\begin{equation}
\begin{aligned}
&d_j
=
\operatorname{atan2}\!\left(
\sin(\hat a_j-a_j),\cos(\hat a_j-a_j)
\right),\\
&e_j(\hat{\mathbf a},\mathbf a)
=\frac{d_j^2}{\sigma_j^2},
\ j\in\{12,13,14\}.
\end{aligned}
\end{equation}
We treat root orientation as a group: selecting any of its roll, pitch, or yaw channels selects all three. We compare their rotation matrices following:
\begin{equation}
\begin{aligned}
\mathbf R(\mathbf a)&=
\mathbf R_z(a_{20})
\mathbf R_y(a_{19})
\mathbf R_x(a_{18}),\\
e_{\mathrm{rot}}(\hat{\mathbf a},\mathbf a)
&=
\frac{
\|\mathbf R(\hat{\mathbf a})-\mathbf R(\mathbf a)\|_F^2
}{
2\sum_{r=18}^{20}\sigma_r^2
},
\end{aligned}
\end{equation}
We assign this group error to each of the three root-orientation channels:
\begin{equation}
e_j(\hat{\mathbf a},\mathbf a)
=
e_{\mathrm{rot}}(\hat{\mathbf a},\mathbf a),
\qquad j\in\{18,19,20\}.
\end{equation}
These channel-wise errors form the error vector \(e(\hat{\mathbf a},\mathbf a)\) used for masked averaging in Eq. (5). 

\paragraph{Training Details}
After completing 10,000 student distillation updates, we freeze the motion generator and train ARA for 40,000 updates. The VAE remains frozen throughout ARA training.
The training proposal set contains two sampled proposals for each of the 9,974 source motions in the HumanML3D training split. Selection masks and target perturbations are resampled during training. We set the branch weights to \(\lambda_{\mathrm{GT}}=0.5\), \(\lambda_{\mathrm{ID}}=0.25\), and \(\lambda_{\mathrm{Edit}}=0.25\), and set \(\lambda_q=0.1\).
We optimize ARA using AdamW with a batch size of 1,536 and weight decay of \(10^{-4}\). The learning rate increases to \(2\times10^{-4}\) over the first 250 updates, then follows a cosine decay schedule to \(4\times10^{-5}\), or \(20\%\) of its peak value.

\subsection{Sampling Selection Masks and Edit Targets}
We construct selection masks and target attribute sequences separately for the three training objectives introduced in Section 3.3 of the main paper: paired-target supervision (GT), identity supervision (ID), and selective editing (Edit).

\paragraph{Paired-Target Supervision (GT)} 
For paired-target supervision, we use the attributes of the paired ground-truth motion as targets and select all 21 channels:
\begin{equation}
\mathbf A_{\mathrm{GT}}
=
\operatorname{WGAR}(\mathbf X_{\mathrm{GT}}),
\qquad
\mathbf M_{\mathrm{GT}}=\mathbf 1.
\end{equation}

\paragraph{Identity Supervision (ID)} 
For identity supervision, the targets equal the base-motion attributes, \(\mathbf A_{\mathrm{ID}}=\mathbf A_b\). We vary the selection mask to train ARA to preserve the base motion under different attribute-selection patterns.

We sample each mask in three steps. First, we organize the 21 channels into eight groups: left arm, left leg, right arm, right leg, torso, root horizontal displacement, root height, and root orientation. We then uniformly sample one of the 21 channels as the \textbf{anchor channel}; the group containing it is the \textbf{anchor group}.

Second, we sample one of the following six mask constructions using the probabilities in Table~\ref{tab:mask_construction}. Every construction includes the anchor channel.
\begin{itemize}
\item \textit{Single anchor channel:} select only the anchor channel.
\item \textit{Complete anchor group:} select all channels in the anchor group.
\item \textit{Multiple complete groups:} select the entire anchor group, then independently select each remaining group with probability \(0.5\), including all channels in each selected group.
\item \textit{Sparse channels across groups:} include the anchor group and one uniformly sampled different group, then independently include each remaining group with probability \(0.5\). Select exactly one channel from each included group, using the anchor channel for the anchor group and uniformly sampling a channel from every other included group.
\item \textit{Arbitrary channel subset:} uniformly sample a subset size from 2 to 20, then select that many channels, always including the anchor.
\item \textit{All channels:} select all 21 channels.
\end{itemize}
Finally, we enforce the root-orientation grouping rule: if any roll, pitch, or yaw channel is selected, all three are included in the final mask. This final expansion applies to every mask construction.

\begin{table}[t]
\centering
\caption{Mask-sampling distribution for identity supervision and selective editing.}
\label{tab:mask_construction}
\small
\begin{tabular}{@{}lr@{}}
\toprule
Mask construction & Probability \\
\midrule
Single anchor channel        & 0.25 \\
Complete anchor group        & 0.25 \\
Multiple complete groups     & 0.10 \\
Sparse channels across groups & 0.10 \\
Arbitrary channel subset     & 0.10 \\
All channels                 & 0.20 \\
\bottomrule
\end{tabular}
\end{table}

\paragraph{Selective Editing (Edit)}
For selective editing, we sample \(\mathbf M_{\mathrm{Edit}}\) using the same procedure as for identity supervision.

Since paired edited motions are unavailable, we construct targets by transforming the base-motion attributes:

\begin{equation}
\begin{aligned}
\mathbf A_b&=\operatorname{WGAR}(\mathbf X_b),\\
\mathbf A_{\mathrm{Edit}}
&=
(\mathbf 1-\mathbf M_{\mathrm{Edit}})\odot\mathbf A_b
+
\mathbf M_{\mathrm{Edit}}\odot\mathcal P(\mathbf A_{\mathrm{b}}).
\end{aligned}
\end{equation}
where \(\mathcal P\) applies randomly sampled temporal and attribute-value transformations, and \(\odot\) denotes element-wise multiplication. The channel mask is shared across frames: selected channels receive transformed targets, while unselected channels retain their base values.

\begin{itemize}
\item \textit{Temporal transformations.} We modify the timing of attribute trajectories to create targets in which the same variation occurs earlier, later, faster, or slower, while keeping the sequence duration unchanged. Temporal shifting and scaling are enabled independently, each with probability \(0.35\). The temporal shift is sampled from \([-0.15,0.15]\) times the sequence duration,  and the temporal scale is sampled from \([0.8,1.25]\) around the sequence midpoint. We obtain the transformed values through linear interpolation. Sampling outside the sequence boundaries returns the nearest endpoint value.

\item \textit{Attribute-value transformations.} We also modify attribute values to vary the requested geometry. For each limb or torso channel, we independently apply two transformations, each with probability \(0.35\): scaling deviations from the channel's temporal mean by a factor sampled from \([0.75,1.25]\), and adding a constant offset sampled from \([-0.2,0.2]\). Limb offsets are dimensionless, while torso offsets are expressed in radians.

The two root horizontal displacement channels share a scale sampled from \([0.8,1.2]\). We subtract their respective initial values so that both displacement trajectories start at zero. Root height receives a constant offset sampled from \([-0.1,0.1]\) metres with probability \(0.35\). Root roll and pitch independently receive constant offsets sampled from \([-0.15,0.15]\) radians, each with probability \(0.35\). After temporal transformation, we shift the yaw trajectory so that its initial value matches that of the base motion. All values within the specified ranges are sampled uniformly.

\item \textit{Additional anchor perturbation.} With probability \(0.5\), we further perturb the anchor channel to strengthen the edit request. The perturbation magnitude is sampled uniformly from \([0.04,0.20]\), with positive and negative signs chosen with equal probability. For root horizontal displacement and yaw, the offset increases linearly from zero at the first frame to the sampled signed value at the last frame. For all other channels, the offset is constant throughout the sequence. Each offset is expressed in the corresponding channel's native units.
\end{itemize}
We resample masks and transformations during training and exclude padded frames from supervision. The resulting targets serve as soft optimization objectives rather than hard constraints.

\section{Sensitivity to Preservation Thresholds}
\label{sec:supp-asr-analysis}
In the main paper, we evaluate motion editing in terms of both target adherence and source preservation. \textbf{Non-target Attribute Drift (NAD)} measures the normalized deviation of unselected attributes from the source motion, while \textbf{Non-target Joint Drift (NJD)} measures joint-space RMSE over unedited body parts. Lower values indicate better performance for all three metrics.

We report the \textbf{Attribute Success Rate (ASR)} using preservation thresholds of NAD \(\leq 0.30\) and NJD \(\leq 3^\circ\). To examine how these choices affect ASR, we vary the NAD threshold from \(0.10\) to \(0.50\) and the NJD threshold from \(1^\circ\) to \(5^\circ\) in separate sweeps. Smaller thresholds impose stricter preservation requirements. We reuse the same predictions and metric definitions as in the main paper, changing only the threshold being examined. We compare AGLR with other motion editors and analyze the contributions of its training components.

\subsection{Comparison with Motion Editors}
\begin{figure}[!t]
\centering
\includegraphics[width=\linewidth]{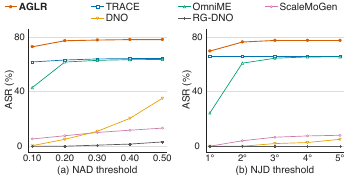}
\caption{\textbf{Editing success under different preservation thresholds.} Attribute Success Rate (ASR) is computed from the same predictions while varying the NAD threshold (left) or NJD threshold (right). }
\label{fig:supp-asr-methods}
\end{figure}
Figure~\ref{fig:supp-asr-methods} shows that AGLR achieves the highest ASR throughout both threshold ranges. TRACE's ASR remains nearly constant as the NJD threshold varies, consistent with its explicit projection of protected joints back to their source configurations. Its lower ASR suggests that preserving these joints alone is insufficient when the edited motion does not closely match the requested targets. OmniME benefits most when the strictest preservation thresholds are relaxed, suggesting that some of its edits achieve the requested changes but also alter unselected attributes or unedited joints. DNO improves as the NAD threshold is relaxed, yet its ASR remains low across the NJD thresholds. RG-DNO and ScaleMoGen also achieve lower ASR than AGLR throughout the evaluated ranges. Overall, AGLR maintains its advantage under both strict and more permissive preservation requirements.

\subsection{Effect of Training Components}
\begin{figure}[!t]
\centering
\includegraphics[width=\linewidth]{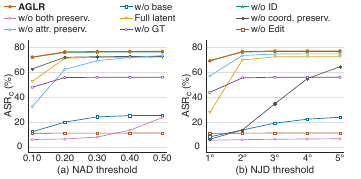}
\caption{\textbf{Contributions of training components under different preservation requirements.} Attribute preservation is particularly beneficial at strict NAD thresholds, while coordinate preservation is particularly beneficial at strict NJD thresholds.}
\label{fig:supp-asr-ablations}
\end{figure}
Figure~\ref{fig:supp-asr-ablations} shows that conditioning on the base motion and predicting a latent residual both improve ASR across the two threshold sweeps. The two preservation objectives provide complementary benefits: attribute preservation is particularly important when unselected attributes must remain close to their source values, while coordinate preservation is particularly important when unedited joints must remain close to their source configurations. Removing both objectives substantially reduces success, even under more permissive thresholds. ID supervision has a smaller effect on these curves, consistent with its primary role in maintaining no-edit consistency, as discussed in the main paper.

\section{Editing Latency Measurement}
\label{sec:F}
We measure cached-refinement latency as the time required for ARA refinement and VAE decoding, starting from a cached base latent corresponding to a 500-frame motion. Measurements are performed on an NVIDIA GeForce RTX 4090D GPU with a batch size of one.
We use three fixed validation motions. For each motion, we run ten warmup requests followed by 30 timed requests, synchronizing the GPU at the timing boundaries. Timing ends when the decoded RPR39 output is available on the GPU. Across the 90 timed requests, the mean latency is 9.103 ms, reported as 9.1 ms in Section 4.4 of the main paper.
The end-to-end editing latency is 13.15 ms and additionally includes source-motion processing and output conversion.

\section{Baseline Adaptation and Training}
\label{sec:supp-control-benchmark}
\subsection{Text-to-Motion Baselines}
\paragraph{MDM}
We adapt MDM~\cite{mdm2023} to RPR39 by modifying its input and output layers to process 39-dimensional robot motion features, while retaining its original diffusion denoising objective. We train the model for 80,000 updates with a global batch size of 128.

\paragraph{ECHO}
For ECHO~\cite{echo2026}, we replace its native motion representation with RPR39 while retaining its temporal U-Net architecture and clean-motion prediction objective. Training uses a 1,000-step diffusion process with a linear noise schedule. The reconstruction loss is the mean squared error over valid frames and motion features, averaged within each sequence and then across the batch. We train ECHO for 50,000 updates with a global batch size of 128, using AdamW with an initial learning rate of \(2\times10^{-4}\) and weight decay of \(10^{-2}\). Gradient norms are clipped at 1, and EMA weights are used for evaluation.

\paragraph{MoMask}
We adapt MoMask~\cite{momask2024} to RPR39 by first training a residual vector-quantized motion tokenizer, then training its masked and residual token predictors using the original cross-entropy objectives. The tokenizer uses a Smooth-\(\ell_1\) motion reconstruction loss, a commitment loss with weight \(0.02\), and an auxiliary G1 FK position reconstruction loss with weight \(0.5\). We train the tokenizer for 80,000 updates with a global batch size of 256, and each token predictor for 60,000 updates with a global batch size of 64. All models use AdamW with a learning rate of \(2\times10^{-4}\), betas \((0.9,0.99)\), and 2,000 warmup updates. Weight decay is zero for the tokenizer and \(10^{-5}\) for the token predictors. The resulting robot-motion tokenizer and token predictors also provide the initialization for MaskControl.

\paragraph{TextOp}
We adapt TextOp~\cite{textop2026} to robot motion while retaining its history-conditioned motion VAE and autoregressive latent diffusion model. At each rollout step, the model conditions on two history frames and predicts eight future frames. For text-only generation, it starts from a fixed G1 neutral pose and uses generated motion as history for subsequent steps. We train the VAE for 40,000 updates using Adam, followed by 60,000 diffusion-model updates using AdamW. Both stages use a global batch size of 1,024 and a learning rate of \(10^{-4}\).

\paragraph{MotionHiFlow}
We adapt MotionHiFlow~\cite{motionhiflow2026} to robot motion while retaining its motion VAE and text-conditioned latent flow model. The VAE compresses the temporal resolution by a factor of four, and the flow model uses 20 sampling steps at inference. We train the VAE for 40,000 updates with a global batch size of 256, followed by 50,000 flow-model updates with a global batch size of 64. Both stages use AdamW with a learning rate of \(2\times10^{-4}\) and zero weight decay. For evaluation, we use the EMA weights of the flow model.

\subsection{Fine-Grained Controllable Generation}
\paragraph{MotionLCM}
We adapt MotionLCM~\cite{motionlcm2024} using a pretrained robot-motion VAE and latent consistency prior. Only the attribute encoder and a ControlNet branch copied from the denoiser are trained; the pretrained components remain frozen. The controller encodes masked attribute targets and selection masks, then injects control features into the denoiser. Training combines clean-latent MSE with Smooth-$\ell_1$ error between decoded WGAR attributes and targets, normalized by training attribute scales and averaged over selected channels and valid frames. Angular differences are wrapped to $[-\pi,\pi]$. Gradients pass through the frozen decoder to train the controller.

We train for 10,000 updates with a global batch size of 192, using AdamW with a learning rate of \(10^{-4}\) and zero weight decay. The learning rate uses a 1,000-update warmup followed by cosine decay. Gradient norms are clipped at 1, and text conditioning is dropped with probability \(0.1\). Inference uses four consistency steps followed by one decoding pass.

\paragraph{OmniControl}
We initialize OmniControl~\cite{omnicontrol2024} from the robot-adapted MDM and jointly fine-tune its backbone and control branch. Analytic guidance differentiates WGAR target errors through G1 forward kinematics, using training-scale normalization and wrapped angular differences on selected channels and valid frames. The denoising loss is RPR39 MSE against the clean training motion over valid frames.

During training, guidance uses 20 updates per diffusion level and 100 at the final level. We fine-tune for 10,000 updates with a global batch size of 192, using AdamW with a learning rate of \(10^{-5}\) and zero weight decay. In the main comparison, inference uses 50 diffusion steps with one guidance update at every step, including the final step.

\paragraph{MaskControl}
We adapt MaskControl~\cite{maskcontrol2025} using the pretrained MoMask tokenizer and token model. A control Transformer copied from the token model injects control features through zero-initialized projections. Its 63-dimensional input contains 21 masked attribute targets, 21 target–current attribute differences, and 21 selection-mask entries.

Training weights token cross-entropy and the attribute-control loss equally. The cross-entropy term supervises ground-truth motion tokens at valid positions. The attribute term averages absolute WGAR target errors over selected channels and valid frames, after training-scale normalization and wrapping angular differences to $[-\pi,\pi]$. To backpropagate this error, we decode probability-weighted token embeddings through the frozen tokenizer. We train for 15,000 updates with a global batch size of 896, using AdamW with a learning rate of \(2\times10^{-4}\) and weight decay of \(10^{-5}\).

Inference uses 18 token-unmasking steps. At the designated refinement stage, we perform five updates to the token logits to reduce attribute error, keeping model weights fixed. These updates use AdamW with a learning rate of \(0.06\), betas \((0.5,0.9)\), and weight decay of \(10^{-6}\).

\paragraph{SFControl}
SFControl~\cite{sfcontrol2025} uses two temporal U-Nets with adaptive group normalization. The first completes 21-channel attribute trajectories from sparse conditions. The second takes these completed trajectories as conditions and predicts a 60-channel state comprising 21 attributes and 39 motion coordinates. Both stages use clean-state MSE. The second also minimizes squared WGAR target errors computed from the predicted motion, normalized by training attribute scales and averaged over selected channels and valid frames, with wrapped angular differences. This loss weight increases to 1 over the first 1,000 updates. We train each stage for 20,000 updates, with global batch sizes of 1,536 and 320, respectively. Both use AdamW with a learning rate of \(10^{-4}\), weight decay of \(0.01\), gradient-norm clipping at 1, and EMA decay of \(0.9999\). Inference uses 50 diffusion steps per stage with a guidance scale of \(2.5\).

\paragraph{TLControl}
TLControl~\cite{tlcontrol2024} first trains a robot-motion VQ-VAE with six body-part groups: root, torso, left/right arms, and left/right legs. A Transformer then predicts motion tokens from masked attribute trajectories and text. The tokenizer minimizes RPR39 reconstruction MSE plus a commitment loss weighted by \(10^{-3}\). The Transformer minimizes token cross-entropy plus decoded-motion MSE weighted by \(10^{-3}\), with the tokenizer frozen. Each stage is trained for 40,000 updates, with global batch sizes of 128 for the tokenizer and 64 for the Transformer. Both use AdamW with a learning rate of \(2\times10^{-4}\), betas \((0.9,0.99)\), and weight decay of \(0.01\). Attribute conditions are dropped with probability \(0.1\) to include text-only training examples.

At inference, the highest-probability tokens initialize the latent representation, which is then optimized continuously. We use at most five L-BFGS iterations with strong-Wolfe line search, a learning rate of \(0.1\), and a history size of 200. The optimization objective is \(1{,}000\) times the sum of the Smooth-$\ell_1$ error on training-scale-normalized WGAR differences, averaged over selected channels and valid frames with angular differences wrapped to $[-\pi,\pi]$ and the G1 foot-skid loss. Additional evaluations required by line search are not counted as iterations.

\paragraph{ProjFlow}
ProjFlow~\cite{projflow2026} uses an RPR39 flow prior trained along the interpolation path \(\mathbf X_\tau=(1-\tau)\boldsymbol\epsilon+\tau\mathbf X_{\mathrm{GT}}\). It predicts the target velocity \(\mathbf X_{\mathrm{GT}}-\boldsymbol\epsilon\) using MSE over valid entries. We train the prior for 50,000 updates with a global batch size of 192, using AdamW with a learning rate of \(2\times10^{-4}\) and weight decay of \(0.01\).

Because WGAR depends on nonlinear FK, we numerically project the predicted clean motion toward the requested attributes. Each projection computes the selected-attribute error gradient, rescales it using the G1 chain-based kinematic metric, and estimates attribute changes along the resulting direction. A line search then tests four progressively smaller step sizes. Inference uses 50 flow steps with a guidance scale of 3, two projection iterations at each non-final step, and eight at the final step. We retain the method's endpoint recomposition and noise-mixing procedure.


\subsection{Source-Conditioned Motion Editing}
\paragraph{Training Pairs for Supervised Editors}
We construct source–target pairs for TRACE~\cite{trace2026} and OmniME~\cite{omnime2026} by numerically editing training motions to match WGAR requests. The solver optimizes a correction in standardized motion space using 120 Adam steps with a learning rate of \(0.025\). Its objective combines selected-attribute error, unselected-attribute error, correction magnitude, and second-difference smoothness, weighted by \(1\), \(0.15\), \(0.01\), and \(0.02\), respectively. Projection restores protected coordinates from the source and enforces joint limits.
Requests use constant, tapered, or sign-changing offsets, with identity requests sampled with probability \(0.25\). We retain all identity pairs and edited pairs whose normalized selected-attribute RMSE is at most \(0.15\). This produces 31,264 training pairs and 256 separately generated validation pairs.

\paragraph{TRACE}
TRACE~\cite{trace2026} retains source-conditioned denoising, edit-scope conditioning, event supervision, and projection of protected coordinates. The request encoder receives masked targets, target--source differences, and masks. Its motion network is initialized from scratch, with frozen Qwen3-Embedding-4B text features. The loss combines edited-coordinate MSE, first-difference MSE, selected-attribute squared error, and event classification with weights $1$, $0.2$, $0.5$, and $0.05$.  Attribute error is computed from WGAR after restoring protected coordinates, using training-scale normalization, wrapped angular differences, and averaging over selected channels and valid frames. Binary and categorical event losses use valid temporal tokens. Training uses AdamW at $10^{-4}$, a 10,000-update cap, global batch 2,048, and EMA decay 0.99. Inference uses 50 DDIM steps and guidance scale 1.5. Protected coordinates are restored at every step and exactly copied in the output.

\paragraph{OmniME}
OmniME~\cite{omnime2026} retains source-conditioned diffusion and adds a framewise 63-channel request encoder to its source tokens. Numerical requests also generate deterministic edit instructions encoded by frozen CLIP ViT-L/14. The eight-layer denoiser has width 512, eight heads, and a four-layer condition encoder. Training uses
\begin{equation}
\begin{aligned}
 \mathcal L_{\mathrm{OmniME}}={}&\mathcal L_{\mathrm{rec}}+\mathcal L_{\mathrm{retro}}+0.05\mathcal L_{\mathrm{cls}}\\
 &+0.2\mathcal L_{\mathrm{pres}}+0.01\mathcal L_{\mathrm{trip}}.
\end{aligned}
\end{equation}
$\mathcal L_{\mathrm{rec}}$ equally averages reconstruction MSE over five RPR39 groups: horizontal increments, height, yaw increment, 6D rotation, and joints. $\mathcal L_{\mathrm{retro}}$ averages three intermediate predictions' motion MSE. $\mathcal L_{\mathrm{cls}}$ classifies three bins of FK-derived source--target similarity. $\mathcal L_{\mathrm{pres}}$ sums squared errors against the edited training target over 39 features and averages valid request-conditioned frames. $\mathcal L_{\mathrm{trip}}$ uses semantic triplets with margin 0.2. Training drops both source and request for 10\% of samples and retains only the source for another 10\%. AdamW uses $10^{-4}$, weight decay 0.01, 500 warmup updates, gradient clipping at 1, and EMA decay 0.99, with global batch 256 and a 31,000-update cap. Inference uses 50 DDIM steps with source and request guidance scales of 2, without output projection.

\paragraph{ScaleMoGen}
ScaleMoGen~\cite{scalemogen2026} replaces human skeleton pooling with G1 joint groups while retaining multiscale binary quantization and coarse-to-fine token prediction. Its robot tokenizer combines smooth-$\ell_1$ reconstruction, velocity $\ell_1$, local-FK $\ell_1$, and commitment loss with weights $1$, $0.5$, $0.5$, and $0.02$. The predictor uses binary cross-entropy over valid multiscale tokens, with masked attributes and masks encoded as temporal condition tokens alongside text. The tokenizer uses 12,000 updates with a global batch size of 32, and the predictor uses 24,000 updates with a global batch size of 24. AdamW uses $2\times10^{-4}$ and weight decay 0.01. Editing extracts source tokens and their confidence scores, then applies the original confidence-difference sampling rule only to the requested body groups. The editing threshold is 0.1 and guidance scale is 3. The edited tokens are decoded directly into motion.

\paragraph{DNO}
DNO~\cite{dno2024} freezes the RPR39 MDM and optimizes sampling noise. Source editing begins with 50 DDIM steps and 500 noise-fitting updates against valid-frame source-motion MSE. Editing uses 300 updates to minimize absolute WGAR target errors, normalized by training attribute scales and averaged over selected channels and valid frames, with wrapped angular differences. Each differentiable sampling pass uses ten DDIM steps. The objective includes the original multiscale noise-decorrelation penalty with weight 1,000. Editing also penalizes the noise distance from the fitted source with weight 0.002, while source fitting omits this distance term. Adam uses learning rate 0.05, a 50-update warmup, cosine decay, and per-sample gradient normalization. Noise penalties use valid sequence lengths, and the output is not projected onto protected source coordinates.

\paragraph{RG-DNO}
RG-DNO~\cite{rgdno2026} uses the RPR39 MDM with 50-step sampling and an attribute-based training-motion retrieval index. After 500 source-fitting updates, it performs 100 target-optimization updates on the source noise, retrieves a reference using selected attributes at 64 sampled frames, and fits and adapts that reference for 100 updates each. It evaluates 255 eight-segment noise-mixture candidates using attribute, jitter, noise-regularity, and foot-skating rewards. The selected mixture receives 100 soft-mask updates followed by 400 final noise updates. Initial stages use learning rate 0.05 and final refinement uses 0.02. Retrieval uses training motions of sufficient valid length. The final optimized noise produces the edited motion through the diffusion sampler.

\section{Limitations}
\label{sec:supp-limitations}
AGLR is trained and evaluated on motion sequences of up to 500 frames at 50 Hz, or approximately 10 seconds. Editing minute-long motions has not yet been evaluated. Because the adapter processes the entire target sequence at once, its cached single-pass refinement does not directly support streaming edits. Extending the method to longer motions will require longer temporal context or a windowed editing strategy that preserves continuity across window boundaries.

\begin{figure*}[!tbp]
\centering
\includegraphics[width=\textwidth]{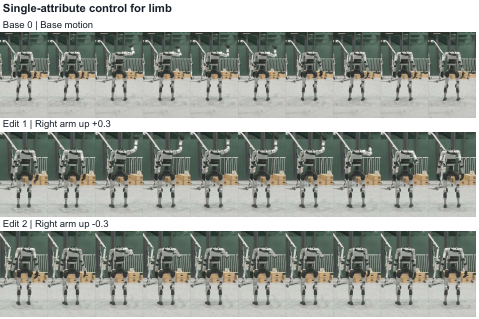}
\caption{\textbf{Right-arm Up control produces opposite vertical adjustments.} The base execution is compared with $+0.3$ and $-0.3$ offsets to the right-arm Up attribute. Offsets are dimensionless changes in normalized limb extent.}
\label{fig:single_attribute_arm_up}
\end{figure*}
  
\begin{figure*}[!tbp]
\centering
\includegraphics[width=\textwidth]{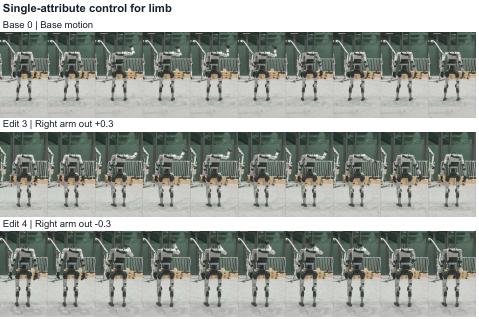}
\caption{\textbf{Right-arm Out control adjusts lateral extension.} Relative to Base~0, offsets of $+0.3$ and $-0.3$ request outward and inward changes, respectively.}
\label{fig:single_attribute_arm_out}
\end{figure*}

\begin{figure*}[!tbp]
\centering
\includegraphics[width=\textwidth]{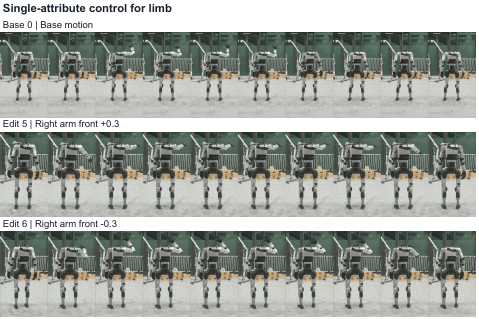}
\caption{\textbf{Right-arm Front control changes forward extension.} The edited executions use Front offsets of $+0.3$ and $-0.3$ relative to Base~0.}
\label{fig:single_attribute_arm_forward}
\end{figure*}

\begin{figure*}[!tbp]
\centering
\includegraphics[width=\textwidth]{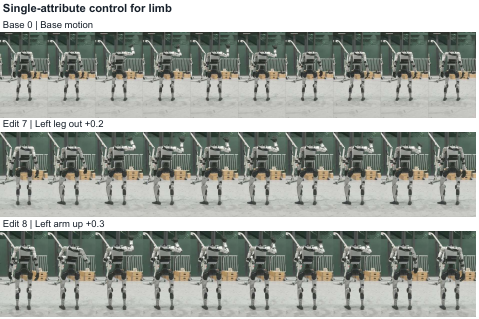}
\captionsetup{skip=4pt}
\caption{\textbf{Single-attribute edits control left-leg and left-arm geometry.} From top to bottom: Base~0, left-leg Out $+0.2$, and left-arm Up $+0.3$. The edits independently increase pelvis-relative lateral leg extent and torso-relative arm height. Offsets are dimensionless WGAR changes. Snapshots progress from left to right.}
\label{fig:single_attribute_left_limbs}
\end{figure*}

\begin{figure*}[!tbp]
\centering
\includegraphics[width=\textwidth]{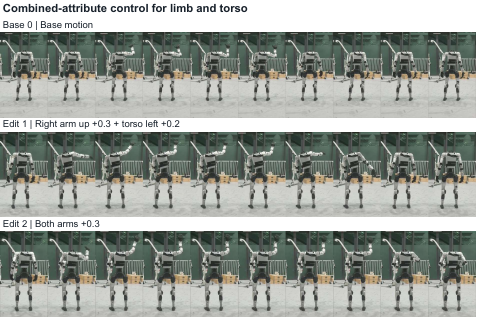}
\caption{\textbf{A single request can combine controls across body parts.} The first edit combines right-arm Up $+0.3$ and torso-left $0.2$. The second increases Up by $0.3$ on both arms (labeled ``Both arms $+0.3$''). Limb offsets and torso-angle adjustments use dimensionless units and radians, respectively.}
\label{fig:combined_attributes_arm}
\end{figure*}

\begin{figure*}[!tbp]
\centering
\includegraphics[width=\textwidth]{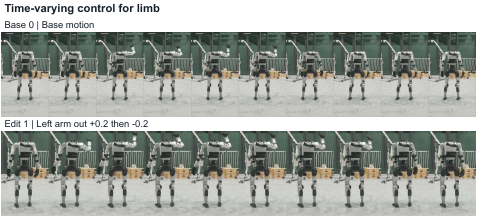}
\caption{\textbf{A time-varying request reverses the direction of a limb edit.} Left-arm Out is first increased by $0.2$ and then decreased by $0.2$ relative to Base~0. The sequence illustrates temporal control within one selected attribute channel.}
\label{fig:time_varying_attributes}
\end{figure*}

\begin{figure*}[!tbp]
\centering
\includegraphics[width=\textwidth]{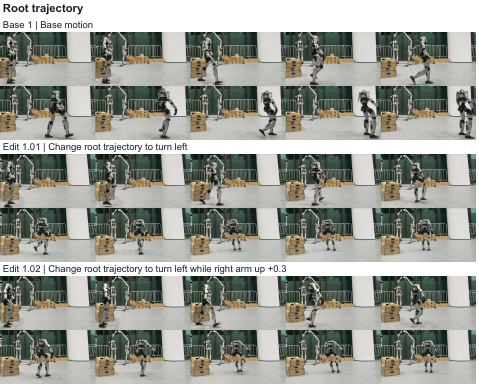}
\caption{\textbf{Root trajectory edits compose with local limb control.} Base~1 is compared with a left-turn request and a left-turn request combined with right-arm Up $+0.3$. Each labeled example is displayed as two rows of successive execution snapshots.}
\label{fig:root_trajectory}
\end{figure*}

\begin{figure*}[!tbp]
\centering
\includegraphics[width=\textwidth]{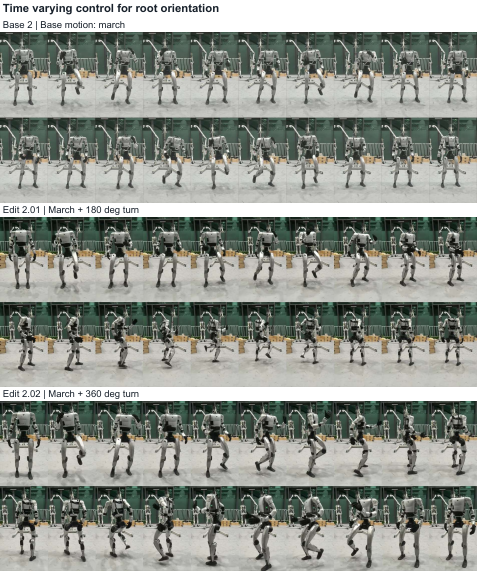}
\caption{\textbf{Time-varying heading targets produce turns during marching.} The base marching execution is compared with $180^{\circ}$ and $360^{\circ}$ turn requests. Successive poses show the progression of each turn.}
\label{fig:root_orientation}
\end{figure*}

\end{document}